\PassOptionsToPackage{usenames,dvipsnames}{xcolor}
\documentclass[10pt,twocolumn,letterpaper]{article}

\usepackage{cvpr}              

\usepackage{graphicx}
\usepackage{amsmath}
\usepackage{amssymb}

\usepackage{amsmath,amsfonts,bm}

\def\eqref#1{Eq.~(\ref{#1})}

\def\1{\bm{1}}

\DeclareMathAlphabet{\mathsfit}{\encodingdefault}{\sfdefault}{m}{sl}
\SetMathAlphabet{\mathsfit}{bold}{\encodingdefault}{\sfdefault}{bx}{n}

\usepackage{url}
\usepackage{bbding}
\usepackage{booktabs}
\usepackage{comment}
\usepackage{enumitem}
\usepackage{wrapfig}
\usepackage{xspace}
\usepackage{xargs}                      
\usepackage{graphicx}
\usepackage{multirow}
\usepackage{caption}
\usepackage{subcaption}
\usepackage{xcolor, soul}
\sethlcolor{green}
\usepackage{array}

\newcommand{\xfill}[2][.7ex]{{%
  \dimen0=#2\advance\dimen0 by #1
  \mbox{}\leaders\hrule height \dimen0 depth -#1\hfill%
}}

\newcommandx{\unsure}[2][1=]{\todo[linecolor=red,backgroundcolor=red!25,bordercolor=red,#1]{#2}}
\newcommandx{\change}[2][1=]{\todo[linecolor=blue,backgroundcolor=blue!25,bordercolor=blue,#1]{#2}}
\newcommandx{\info}[2][1=]{\todo[linecolor=OliveGreen,backgroundcolor=OliveGreen!25,bordercolor=OliveGreen,#1]{#2}}
\newcommandx{\improvement}[2][1=]{\todo[linecolor=Plum,backgroundcolor=Plum!25,bordercolor=Plum,#1]{#2}}
\newcommandx{\thiswillnotshow}[2][1=]{\todo[disable,#1]{#2}}

\newcommand{\upg}[1]{{\scriptsize \color{ForestGreen}$\uparrow$#1}}
\newcommand{\downr}[1]{{\scriptsize \color{BrickRed}$\downarrow$#1}} 

\newcommand{\todo}[1]{\textcolor{red}{#1}}

\usepackage[pagebackref,breaklinks,colorlinks]{hyperref}

\usepackage[capitalize]{cleveref}
\crefname{section}{Sec.}{Secs.}
\Crefname{section}{Section}{Sections}
\Crefname{table}{Table}{Tables}
\crefname{table}{Tab.}{Tabs.}

\def\cvprPaperID{5273} 
\def\confName{CVPR}
\def\confYear{2024}

\newcommand{\bx}{\boldsymbol{x}}

\newcommand{\btheta}{\boldsymbol{\theta}}
\newcommand{\bmu}{\boldsymbol{\mu}}
\newcommand{\bkappa}{\boldsymbol{\kappa}}
\newcommand{\bt}{\boldsymbol{t}}
\newcommand{\by}{\boldsymbol{y}}

\newcommand{\bX}{\mathcal{X}}

\newcommand{\bY}{\mathbf{Y}}
\newcommand{\bT}{\mathcal{T}}

\newcommand{\bD}{\mathbf{X}}
\newcommand{\bset}{\mathbf{D}}
\newcommand{\bsetU}{\mathbf{D}_{\text{uncertain}}}
\newcommand{\bDC}{\bD_{\text{confident}}}
\newcommand{\bDU}{\bD_{\text{uncertain}}}
\newcommand{\bDyR}{\bDy_{\text{refined}}}

\newcommand{\bDy}{\bset}
\newcommand{\bDyU}{\bsetU}

\newcommand{\phit}{\phi_t}
\newcommand{\phiv}{\phi_x}

\newcommand{\HH}{\mathbb{H}}
\newcommand{\bYu}{\mathbf{Y}_{\text{uncertain}}}

\begin{document}
\title{Zero-shot Retrieval:\\Augmenting Pre-trained Models with Search Engines}

\author{
Hamed Damirchi$^1$
\and
Cristian Rodríguez-Opazo$^1$
\and
Ehsan Abbasnejad$^1$
\and
Damien Teney$^2$\\
\vspace{-9mm}
\and
Javen Qinfeng Shi$^1$
\and
Stephen Gould$^3$
\and
Anton van den Hengel$^{1,4}$\\
\vspace{-14mm}
\and
$^1$Australian Institute for Machine Learning, University of Adelaide,
\and
\vspace{-8mm}
$^2$Idiap Research Institute,
$^3$Australian National University,
$^4$Amazon
}
\maketitle

\begin{abstract}
    Large pre-trained models can dramatically reduce the amount of task-specific data required to solve a problem, but they often fail to capture domain-specific nuances out of the box. The Web likely contains the information necessary to excel on any specific application, but identifying the right data a priori is challenging.
    This paper shows how to leverage recent advances in NLP and multi-modal learning to augment a pre-trained model with search engine retrieval.
    We propose to retrieve useful data from the Web at test time
    based on test cases that the model is uncertain about.
    Different from existing retrieval-augmented approaches, we then update the model to address this underlying uncertainty.
    We demonstrate substantial improvements in zero-shot performance, e.g. a remarkable increase of 15 percentage points in accuracy on the Stanford Cars and Flowers datasets.
    We also present extensive experiments that explore the impact of noisy retrieval and different learning strategies.
\end{abstract}

\section{Introduction}

The World Wide Web is the source of most training data for today's largest models.
The abundance and diversity of the data available online has played a pivotal role in developing state-of-the-art models for vision, natural language processing (NLP), and multi-modal tasks
(\eg DINO~\cite{dino}, GPT~\cite{brown2020language} LLama~\cite{llama}), Flamingo~\cite{flamingo}).
Large pre-trained models have shown remarkable versatility across domains, and settings,
but the models' breadth often hide a number of limitations.
Data quality and curation remains important~\cite{fang2022data,nguyen2022quality},
and increasing dataset sizes only masks the models' failure to generalize to
out of distribution data~\cite{id_ood}.

\begin{figure*} 
\centering
\vspace{-2mm}
\includegraphics[width=0.87\textwidth]{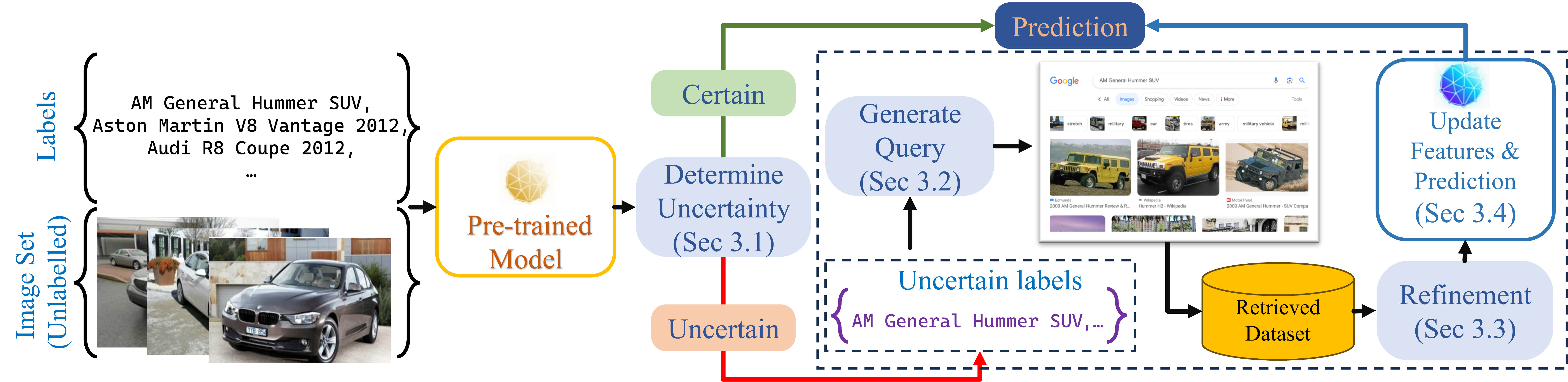}
\vspace{-1mm}
\caption{We propose a method to enrich a pre-trained visual recognition model using web search.
At test time, we determine the model's uncertainty about the given input and candidate labels.
We then issue a text query to acquire additional data, which we use to update the model.
The predictions of the original and updated models are compared to assess the information gained and generate the final output.}
\label{fig:model}
\vspace{-5mm}
\end{figure*}

One  popular method for addressing some of the limitations of large pre-trained models is retrieval-augmented inference~\cite{teney2019actively,realm,seeker_retrieval}.
This method retrieves instances at test time to help with generalisation. Existing methods rely on a supplementary (\eg up-to-date or domain-specific) dataset, from which the model can retrieve information to supplement its pretrained knowledge. Identifying relevant instances is challenging, these methods incur increased computational cost at test time, and the limitations related to finite datasets remain~\cite{wang2020survey,xiong2021answering}.

In contrast to the fixed static datasets of machine learning, humans continuously update their knowledge using new information, often acquired by querying the Web through search engines.
The idea of connecting learning algorithms with the web and search engines is not new but is challenging to implement effectively~\cite{carlson2010,chen2013neil,chen2015webly,shen2018bootstrapping,teney2019actively}. We present here a novel approach that exploits  recent advances in NLP and multimodal learning~\cite{dreamfields,girdhar2023imagebind,clipsonic} to significantly improve upon the state-of-the-art.




This paper explores the synergy between pre-trained models, which capture rich concept priors, and search engines, which are designed to retrieve diverse and representative information based on text queries. We propose a methodology in which a small set of retrieved examples from the search engine is used for fine-tuning the pre-trained model for improved predictions~(see Figure~\ref{fig:model}. Since fine-tuning generally entails making only minimal updates to model parameters~\cite{lora}, we proceed to train a compact model tailored for the task using this retrieved set.  

Contemporary to our efforts, \cite{internetexplorer} recently proposed Internet Explorer, a method for representation learning that also involves an active exploration of the web. Their approach, formulated as reinforcement learning, samples concepts from ConceptNet~\cite{speer2017conceptnet} to uncover potentially-relevant images.
Our approach aims to be much more frugal and efficient in retrieving data.
We use a more \emph{directed} search strategy, and the process is only triggered when the pre-trained model shows high uncertainty.

Our approach distinguishes itself from traditional active learning~\cite{settles2010active,parvaneh2022active} where the objective is to minimize the number of labels requested from annotators, given a fixed set of candidate examples.
In contrast, our approach seeks new instances by formulating a textual search query. It then uses them as noisy, weakly-supervised examples to improve the model.





Technically, we identify concepts whose representation in the pretrained model (CLIP) could be enhanced, by examining the distribution of image projections onto the unit hypersphere.
Interestingly, we observe that instances with low predictive entropy  (\ie deemed certain) remain largely unchanged after fine-tuning with retrieved images. This suggests that the pre-trained model may require adjustments in its conceptual decision boundaries to better handle nuanced instances and accurately classify the desired classes.
We discover that the underlying representations allow us to establish a meaningful correlation between the zero-shot task--the unlabeled images and the list of class labels--and the improvement achieved from employing our approach. We further, found if the task is to benefit from our approach, the best margin of improvement is achieved when the class name (as provided by the dataset) is used as the search query rather than more complicated alternatives such as descriptions or instance-level queries from image captioning~\cite{li2022blip}. {This highlights the efficacy of contemporary search engines in aggregating related concepts and the generic class-level nature of features in datasets. }

In summary, the contributions of this paper are:
\begin{itemize}[leftmargin=*,topsep=1pt,itemsep=2ex,parsep=-1ex]     
    \item We propose a novel approach for enriching a pre-trained visual recognition model through test-time access to web search with no need for additional labels or manual inputs. We show this approach can easily be integrated into the current machine learning pipelines and leads to improved performance.

    \item We implement the method on top of CLIP and derive a measure of uncertainty based on the classification entropy to make the method efficient in the amount of retrieved data. Technically, we use the projection of image embeddings onto a hypersphere and characterize the distribution of underlying concepts using a mixture of von Mises-Fisher distributions.

    \item We conduct extensive experiments to explore various implementation choices. We demonstrate significant improvements on the Stanford Cars~\cite{krause2013collecting} and Flowers~\cite{nilsback2008automated} datasets with remarkable improvements of over 15 percentage points in accuracy. We demonstrate a strong correlation between the observed improvements and the specific characteristics of the zero-shot task at hand. This correlation allows us to proactively assess whether the proposed search-engine augmented approach is advantageous for the task. In cases where our approach proves unsuitable for the task, the acquisition of manually labelled data is a viable alternative. 
\end{itemize}

\section{Related Work}

\noindent\textbf{Using unlabelled or weakly labelled data.}
One area that shares some similarities with the proposed approach is active learning~\cite{settles2010active}. 
In this paper, instead of relying on human experts, the proposed approach leverages the vast knowledge source available on the internet.
The proposed approach also connects to weakly supervised learning, where the aim is to learn from data that is only partially labelled~\cite{zhou2018brief}. In this case, the data collected from the search engine provides weak supervision for enhancing the model's performance. While weakly supervised learning often focuses on improving the quality of the labelling, the proposed approach aims to leverage the search engine as a source of inductive biases to improve the model's predictions.

\noindent\textbf{Cross-Modal retrieval.}
Cross-modal retrieval is another area that is relevant to the proposed approach, where the goal is to retrieve relevant data from different modalities such as text, image, and video~\cite{jiang2017deep}. While cross-modal retrieval typically focuses on retrieving data from a fixed repository, the proposed approach enables real-time, up-to-date, and relevant information to be retrieved from the internet using natural language queries.

\noindent\textbf{Language models.}
The success of the proposed approach also relies on recent advances in language models, such as OpenAI's GPT-3, which enables machines to interact and process natural language text~\cite{brown2020language}. Jointly learning image and text in models such as OpenAI's CLIP~\cite{radford2021learning} or Google's ALIGN~\cite{radford2021learning} have shown promise for improving the performance of visual models. In addition, self-supervised learning and generative modelling have shown promise for learning from noisy labels~\cite{chen2020simple}. By utilizing these technologies, the proposed approach takes a significant leap forward that otherwise would not be possible.

\noindent\textbf{Retrieval-augmented.}
Retrieval-augmented learning offers an exciting avenue for enhancing the capabilities of models in an era where access to vast and dynamic information is paramount. For instance,~\cite{rapre},~\cite{realm},~\cite{seeker_retrieval} learn to leverage large additional data sources for predictions. 
However, that requires using a common embedding space and implementation of indexing mechanisms.

\noindent\textbf{Webly-supervised learning.}
NELL~\cite{carlson2010toward} and NEIL~\cite{chen2013neil} are amongst the first to explore acquiring new concepts and relationships that are periodically being refined with human supervision from the web. 
Ideas from webly-supervised learning, where the focus is on harnessing the vast and noisy data available on the web have lent ideas for large pre-training~\cite{guo2018webly,Guo2017ece,dai2018large}. In~\cite{beser2021zero} and~\cite{li2018learning}  webly supervised zero-shot learning in natural language processing is explored. Further,~\cite{zhang2017webly} demonstrates how web data can enhance fine-grained categorization tasks.~\cite{zang2019webly,zhang2017webly} combines webly supervised and zero-shot learning techniques for fine-grained recognition tasks. Together, these works offer valuable insights into leveraging web data for training machine learning models.  However, using search engines where text is the medium for representation remains under-explored.

\section{Multimodal Pre-trained Models}
\label{sec:pretrained}
A large pre-trained model such as CLIP is trained using a self-supervised objective to align the output (\ie, representations) obtained from both language and image outputs, respectively. The significance of this is that when trained (unsupervised), we can map either textual or visual inputs using their corresponding neural encoders $\phi_t, \phi_x$ to the same semantic space (assuming the outputs are normalized). One way to interpret these models is to consider an image $\bx\in\bX$ and its corresponding textual description $\bt \in\bT$ as $p(\bx,\bt) \propto \exp(\langle \phiv(\bx), \phit(\bt) \rangle).$
Here, $\bX,\bT$ denote the space of image and text collected in a large corpus.
Then, for the downstream classification task with the label set $\bY$, we have:
{\small\begin{align}
    p_{\text{pre}}(\by\mid \bx) =\frac{ \exp\big(\langle \phiv(\bx), \phit(\by)\rangle\big) }{ \sum_{\by'\in\bY} \exp(\langle \phiv(\bx), \phit(\by') \rangle)}, \label{eq:conditional}\,,
\end{align}}
where the labels $\by$ are used in their textual form, benefiting from the inherent capability of such pre-trained models for open-ended problems. The label is predicted as $\by^\star = \arg\max_{\by'\in \bY}~~p_{\text{pre}}\big(\by'\mid \bx\big)$.

\section{Search-engine Augmented Learning}
\label{sec:learning}

Our main goal is to improve the performance of a pre-trained model at inference time, \ie CLIP. In our setting, we are given an (unlabelled) ``target'' dataset $\bD$ of images and a list of potential labels $\bY$. We are interested in training a complementary model on a ``small'' retrieved dataset to predict the target labels.
To build such a dataset, we leverage the vast amount of data on the internet by creating a retrieval mechanism (note there is no curated labelled training set). We propose a retrieval mechanism that considers the uncertainty in the classification of the pre-trained model, queries and downloads images from search engines like Google. Our retrieval mechanism aims to effectively cover the uncertain classes in the target dataset. The retrieved dataset is expected to be noisy, containing unrelated images for certain class names. For example, one can look after images of flowers with the class name `Prince of Wales feathers' and retrieve images of royalty instead of flowers. In light of this, our retrieval mechanism also incorporates a refinement process to clean the dataset before training a classifier.  
We consider the following steps as the general algorithm:
\begin{enumerate}[leftmargin=20pt,topsep=0pt,itemsep=1ex,parsep=-1ex]
\item determine the uncertain instances in the given unlabelled target dataset (see Sec. \ref{sec:uncertain})
\item formulate a query and invoke the search engine to retrieve the relevant images (see Sec. \ref{lab:algo_steps})
\item refine and filter out the unrelated images (see Sec. \ref{sec:refinement})
\item train a small model (\eg a linear probe) for classification (see Sec. \ref{sec:prediction})
\end{enumerate}
Formally, for the set of uncertain labels $\bYu$, \emph{at inference time}, with access to a search engine $q_{\text{se}}$ that allows sampling $n$ most appropriate images given an uncertain class label $\by_r$, we consider
\begin{align}
    \btheta^\star \!=\! \arg\max_{\btheta}\, 
    \frac{1}{n}\!\sum_{\bx_r,\by_r}\! \log\left(p_{\btheta}(\by_r\mid\bx_r)^{\mathbb{I}[f(\bx_r\mid\bD)]}\right),\\
    ~~\bx_r\!\sim\! q_{\text{se}}(\bx_r\mid \by_r,\bD),\, \forall\,\by_r\!\in\! \bYu\,,
\end{align}
where $\btheta$ is a small number of parameters compared to the ones in the pre-trained model (\eg a simple classifier optimized on a frozen CLIP backbone) and $f(\bx_r\mid\bD)$ is a function indicating whether a given sample $\bx_r$ belongs to the same distribution as those in $\bD$ (see Sec. \ref{sec:refinement}). 
The optimal inference-time $\btheta^\star$ are the maximum likelihood parameters for the set  $\{(\bx_r, \by_r)\}$ obtained from the search engine.
Effectively, a search engine enables us to \emph{sample from the distribution} of all images potentially relevant to a query $\by_r$.


\subsection{Uncertainty in Pre-trained models}
\label{sec:uncertain}
To construct the dataset, we focus on the image samples whose label is uncertain for the pre-trained model. Considering the  conditional probability in \eqref{eq:conditional}, we can use the Shannon entropy as a measure of uncertainty, with a  threshold $\tau_H$  to construct a subset of uncertain instances \ie
\begin{align}
    \bDU = \{ \bx ~\big|~ \HH_{\text{pre}}(\by\mid\bx) \geq \tau_H, \forall \bx\in\bD \}\,.
\end{align}
Note that this uncertainty measures the uniformity of predictions for the given labels using the pre-trained model. If the model is not confident, \ie produces a higher score for a class, this uncertainty will be high. We collect samples whose entropy is higher than a threshold needing additional information for prediction. We pick all the class labels in the top-$k$ predictions for the uncertain instances $\bx\in\bDU$ and create an uncertain label set $\bYu$. 

\subsection{Creating a Search Engine-based Dataset}
\label{lab:algo_steps}
With the uncertain label set identified, we can now generate queries to obtain relevant images from the search engine. In this stage, we leverage the uncertain class labels to construct a query for the search engine. We subsequently collect the retrieved images.
We consider the following strategies for search queries for each uncertain class: 
{\small\begin{enumerate}[leftmargin=20pt,topsep=0pt,itemsep=1ex,parsep=-1ex]
\item \textbf{Classnames} ($p^\texttt{cls}$): Directly using the ``\texttt{\{class\_name\}}" which is the name given to the class as per the dataset (\eg CLIP Benchmark \cite{clip_benchmark}). In other words, for a specific uncertain label, we simply search for the class name. 

\item \textbf{Descriptions} ($p^\texttt{desc}$): Employing a LLM, \ie GPT-3, to generate ``\{description\}"s for the class names similar to~\cite{menon2022visual}. For each class, we can generate multiple descriptions. Then, the class names are concatenated with these descriptions, \ie ``\texttt{\{class\_name\} which (is/has/etc) \{description\}\}}" to produce a search query. 

\item \textbf{Captioning} ($p^\texttt{cap}$): Using a captioning method such as BLIP~\cite{li2022blip}. The class name is then concatenated with this caption as ``\texttt{\{class\_name\} \{caption\}}". Note that in this approach, the query is constructed for each instance individually as opposed to the previous two alternatives.
\end{enumerate}}
We can now create a dataset of retrieved images. For instance, for \textbf{Classnames}, we have: 
{\small\begin{align}
\bsetU^{\texttt{cls}} & = \Big\{ (\bx_r, \by_r)\, \Big|\, \bx_r\sim q_{\text{se}}(\bx_r\mid\by_r,\bD)\,,\forall\by_r\in\bYu \Big\}, \nonumber\\
\text{where} ~ &q_{\text{se}}(\bx_r\mid\by_r,\bD) = \sum_{\tilde{\by}_r} p(\bx_r\mid\tilde{\by}_r) p^{\texttt{cls}}(\tilde{\by}_r\mid \by_r, \bD),\nonumber \\ 
& ~~~~~~~ p(\bx_r\mid\tilde{\by}_r)  = \mathbb{I}\left[\bx_r\in\textbf{SearchEngine}(\tilde{\by}_r, n))\right],\!\!\vspace{-5mm}
\end{align}}
and $n$ indicates the number of top images retrieved. Here, $\tilde{\by}_r$ denotes the search query obtained using one of the strategies above (\ie classnames,  similarly for description or captioning). 
We use $\bsetU$ to denote any dataset constructed from the above-mentioned strategies.
\subsection{Refinement}
\label{sec:refinement}
Ideally, the samples in this set should consist of images from a distribution similar to that of the target dataset, yet possessing distinctive features that can enhance performance. However, either due to query ambiguity or dataset specificity, there might be images in the retrieved set $\bDyU$ that do not conform to the distribution of those in  $\bD$. 
\begin{figure}
\centering
\includegraphics[width=0.25\textwidth]{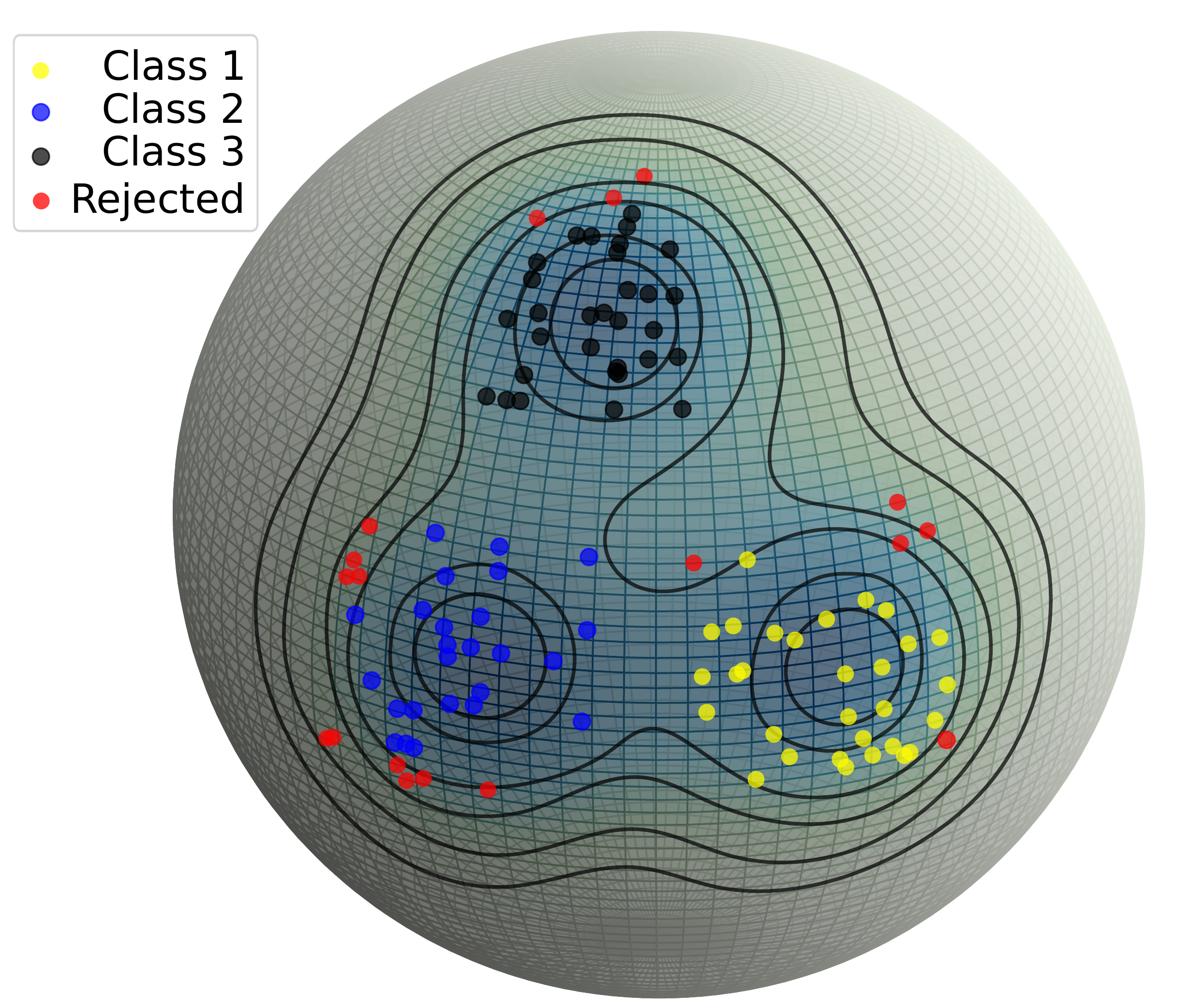}
\caption{\small{Conceptual visualisation of images on the unit hypersphere. Images from three classes are represented by dots, with the colourmap indicating probability density from a mixture of von Mises-Fisher distributions. Threshold lines demonstrate selection boundaries, with images beyond these lines (in red) deemed too distant for selection.}}
\label{fig:refinement_sphere}
\vspace{-5mm}
\end{figure}
 Therefore, a refinement step is imperative to effectively dispose of the potentially noisy samples to create a refined subset $\bDyR$. This necessitates a comparison between the distributions of the retrieved and target datasets.
 Instead of adopting complex density estimation models, such as those necessitating the training of additional neural networks (\eg,~\cite{vae, gade, gade1, diffusion}), we note CLIP's image embeddings, like many other contrastive learning alternatives, are inherently normalised to the unit hypersphere~\cite{clip-understanding} (see Fig.~\ref{fig:refinement_sphere}). Thus, we leverage the mixture of von Mises-Fisher Distributions (MoVMF)~\cite{movmf2005jmlr} to model this hypersphere distribution. The mixture of von Mises-Fisher distributions is a statistical model that combines multiple von Mises-Fisher distributions to represent data distributed on a hypersphere in high-dimensional space, allowing for the modelling of complex directional data patterns.  Once optimised, MoVMF allows us to represent the target distribution as $p(\bx\mid\bD) = \sum_i\pi_i p_{\bmu_i,\bkappa_i}(\bx)$, where $\pi_i$ represents the mixing coefficient, and $\bmu_i$ and $\bkappa_i$ are the parameters of the von Mises-Fisher Distribution. We then utilise the most likely component to assign a hard cluster for each instance.
 Subsequently, we only keep samples that are sufficiently close to the mean of each von Mises-Fisher component, \ie
{\begin{align}
f(\bx_r\mid\bD) &=  \mathbb{I}\big[\langle \phi_x(\bx_r), \bmu_{i^*} \rangle  < \tau_R\big],\\
\quad i^* &= \arg\max_i\, p_{\bmu_i,\bkappa_i}(\bx_r).    
\end{align}}


We explore two alternative approaches (see appendix), but empirical results demonstrate MoVMF outperforms them. It is straightforward to implement and imposes minimal assumptions on the data. 

\subsection{Prediction}
\label{sec:prediction}
In the final step, we train a classifier $p_{\btheta}$ on the refined retrieved samples. However, if there is no information gained (\ie information gain or mutual information of label and additional data is not positive), we fall back to the pre-trained model. In other words, we only use this new model if it leads to a reduced entropy (\ie the confidence in predictions improves):
\begin{equation}
    \by^*=\arg\max_{\by}
    \begin{cases}
      p_{\btheta^\star}(\by\mid\bx), & \text{if}~~\text{IG}(\by\mid\bx,\bDyR)\leq 0  \\ 
      p_{\text{pre}}(\by\mid\bx), & \text{otherwise}
    \end{cases}\,, 
    \label{eq:prediction}
\end{equation}
where $\by^*$ is the predicted class for $\bx$, ${\footnotesize{\text{IG}(\by\mid\bx,\bDyR)=\HH_{\text{pre}}(\by\mid\bx) - \HH_{{\btheta^\star}}(\by\mid\bx,\bDyR)}}$ is the information gain and $\HH_{\btheta^\star}(\by\mid\bx,\bDyR)$ denotes the entropy of the classifier $p_{\btheta^\star}$ (\ie the entropy once the model is updated). 

\section{Experiments}



We employ CLIP~\cite{radford2019language} as the pre-trained model to validate our approach. Our focus primarily lies in a zero-shot scenario, wherein we work under the assumption of not having access to a training dataset. Consequently, our initial comparison revolves around assessing the performance against the baseline zero-shot CLIP. Nevertheless, we also provide supplementary experiments exploring alternative zero or few-shot settings, which can be found in the appendix for a more comprehensive understanding of our approach. It's crucial to note that our methodology refrains from utilizeing any labelled training data except in explicitly specified cases.


\subsection{Datasets and metrics}
We evaluate our method on five popular image classification datasets: Flowers~\cite{nilsback2008automated}, Pets~\cite{parkhi2012cats}, Stanford Cars~\cite{krause2013collecting}, Food~\cite{bossard2014food} and ImageNet~\cite{deng2009imagenet}. These datasets consist of 102, 37, 196, 101 and 1000 classes. Our objective is to retrieve a dataset that matches the distribution of the target dataset, acknowledging the need for CLIP for more information. We evaluate the impact of the retrieved dataset $\bDyU$ by training a linear probe and combining its predictions with those made by CLIP using the accuracy metric. We set $n=100+10$ for the number of instances retrieved from the search engine with the padded amount to account for the failure of the scraper in retrieving all $100$ images.
Moreover, we visually compare the distribution of the target and retrieved dataset using UMAP. We evaluated the overlap between the retrieved and the target dataset, Appendix \ref{app:eval_data_leaks}, our datasets $\bsetU^{\texttt{cls}}$, $\bsetU^{\texttt{cap}}$ and $\bsetU^{\texttt{desc}}$ exhibit negligible overlap with the target testing dataset.

\subsection{What tasks are expected to benefit?}
Before using $\bsetU$ for classifier training, we evaluate the quality of retrieved images by measuring their distance from the dataset in CLIP's feature space which would inform us of any domain shifts between the two sets of images. Additionally, when datasets contain labels with higher specificity, the domain shift between the retrieved images and the dataset is expected to be reduced. So we use the span of the text modality as a proxy for specificity. These values are detailed in Table \ref{tab:featurespace}. The label span is derived using the cosine similarity between the two most distant label names in the CLIP feature space. Similarly, the dataset-to-image similarity is determined using the cosine similarity between the mean vectors of each image set (titled modality similarity in Table \ref{tab:featurespace}). Another similar approach can be taken where instead of the entire image set, images of the same class from the retrieved dataset can be compared to the dataset images (titled class similarity in Table \ref{tab:featurespace}). In terms of the label span, ImageNet obtains the broadest span (lowest similarity) which is expected due to the large number of classes present in this dataset. In contrast, the span of labels in Food101 forms the narrowest cone in the CLIP feature space among all the studied datasets suggesting either low specificity or insufficient training data per category. As for the image similarities in Table \ref{tab:featurespace}, the features of retrieved images for Cars and Flowers are the closest to that of the target dataset. Meanwhile, the noticeable deviation for the same value for Food101 indicates that the retrieved images for this dataset might not be as informative for the classifier due to domain shifts between the dataset and retrieved images, leading to a smaller accuracy improvement.

\begin{table}[h]
    \centering
    \resizebox{\columnwidth}{!}{%
    \begin{tabular}{lccccc}
        \toprule
        & Pets & Flowers & Cars & Food101 & ImageNet \\ \midrule
        Label Span & 0.2885 & 0.2620 & 0.3298 & 0.4496 & 0.1579 \\ 
        Dataset-$\bDyU$ Modality Sim. & 0.9618 & 0.9836 & 0.9889 & 0.8943 & 0.9671 \\
        Dataset-$\bDyU$ Class Sim. & 0.9540 & 0.9590 & 0.9800 & 0.8930 & 0.9350 \\
        \bottomrule
    \end{tabular}}
    \caption{Geometric analysis over the CLIP feature space and the expected improvements. As shown, both label span and dataset similarity for Food101 are substantially different from the others.}
    \label{tab:featurespace}
    \vspace{-2mm}
\end{table}

\subsection{Results and analysis}
\label{results-analysis}

\begin{table}
\centering
\resizebox{0.49\textwidth}{!}{%
\centering
\begin{tabular}{rlllll}
\toprule

                                                                                    & \multicolumn{1}{c}{Flowers} & \multicolumn{1}{c}{Pets} & \multicolumn{1}{c}{Cars} & \multicolumn{1}{c}{Food} & \multicolumn{1}{c}{ImageNet} \\ \cmidrule(lr){2-6}
\cmidrule(lr){2-6}\textit{ZS CLIP-B-16}                                 & 71.15                       & 89.04                    & 64.71                    & 88.73                    & 68.33                        \\
Ours                                                                    & 86.62 \upg{15.47}           & 93.00 \upg{3.96}         & 82.38 \upg{17.67}        & 88.97 \upg{0.24}         & 70.33 \upg{2.00}             \\ \cmidrule(lr){2-6}
LP w/ $\mathbf{T}_r$                                       & 96.41                       & 92.72                    & 85.64                    & 92.08                    & 78.66                        \\ 
\begin{tabular}[c]{@{}c@{}}LP w/ $\bDyU$ + $\mathbf{T}_r$ \end{tabular} & 96.57 \upg{0.16}           & 93.62 \upg{0.9}          & 86.03 \upg{0.39}         & 92.15 \upg{0.07}         & 78.94 \upg{0.28}             \\
 \bottomrule
\end{tabular}
}
\caption{\small{Comparison of training a {LinearProbe} (LP) using (1) the labelled training set ($\mathbf{T}_r$) and (2) the training set ($\mathbf{T}_r$) augmented with the retrieved dataset $\bDyU$.}}
\label{tab:main_table}
\vspace{-3mm}
\end{table}

\begin{table*}
\centering
\resizebox{0.7\textwidth}{!}{%
\begin{tabular}{@{}rccc|ccccc@{}}
\toprule
\multirow{2}{*}{Dataset} & \multicolumn{3}{c|}{Number of Images} & \multicolumn{5}{c}{Accuracy Results (\%)} \\
 & \multicolumn{3}{c|}{\small{(Unrefined / Refined)}} & \multicolumn{5}{c}{\small{(Unrefined / Refined)}} \\
\midrule
 & $\bsetU^{\texttt{cls}}$ & $\bsetU^{\texttt{cap}}$ & $\bsetU^{\texttt{des}}$ & Zero-Shot & $\bsetU^{\texttt{cls}}$ & $\bsetU^{\texttt{cls+cap}}$ & $\bsetU^{\texttt{cls+des}}$ & $\bsetU^{\texttt{cls+cap+des}}$ \\
\midrule
Pets & \begin{tabular}[c]{@{}c@{}}4096\\4011\end{tabular} & \begin{tabular}[c]{@{}c@{}}50653\\50428\end{tabular} & \begin{tabular}[c]{@{}c@{}}23226\\22973\end{tabular} & 89.04 & \begin{tabular}[c]{@{}c@{}}92.61 \upg{3.57}\\92.64 \upg{3.60}\end{tabular} & \begin{tabular}[c]{@{}c@{}}92.45 \upg{3.41}\\92.56 \upg{3.52}\end{tabular} & \begin{tabular}[c]{@{}c@{}}92.59 \upg{3.55}\\92.67 \upg{3.63}\end{tabular} & \begin{tabular}[c]{@{}c@{}}92.78 \upg{3.74}\\\textbf{93.00} \upg{3.96}\end{tabular} \\ 
\midrule
Flowers & \begin{tabular}[c]{@{}c@{}}11197\\10938\end{tabular} & \begin{tabular}[c]{@{}c@{}}27932\\27506\end{tabular} & \begin{tabular}[c]{@{}c@{}}79508\\75111\end{tabular} & 71.15 & \begin{tabular}[c]{@{}c@{}}86.24 \upg{15.09}\\86.32 \upg{15.17}\end{tabular} & \begin{tabular}[c]{@{}c@{}}85.44 \upg{14.29}\\86.27 \upg{15.12}\end{tabular} & \begin{tabular}[c]{@{}c@{}}85.77 \upg{14.62}\\86.09 \upg{14.94}\end{tabular} & \begin{tabular}[c]{@{}c@{}}86.36 \upg{15.21}\\\textbf{86.62} \upg{15.47}\end{tabular} \\
\midrule
Cars & \begin{tabular}[c]{@{}c@{}}21514\\19416\end{tabular} & \begin{tabular}[c]{@{}c@{}}149820\\139458\end{tabular} & \begin{tabular}[c]{@{}c@{}}156491\\120293\end{tabular} & 64.71 & \begin{tabular}[c]{@{}c@{}}80.52 \upg{15.81}\\80.48 \upg{15.77}\end{tabular} & \begin{tabular}[c]{@{}c@{}}80.33 \upg{15.62}\\81.82 \upg{17.11}\end{tabular} & \begin{tabular}[c]{@{}c@{}}80.81 \upg{17.10}\\80.79 \upg{17.08}\end{tabular} & \begin{tabular}[c]{@{}c@{}}82.32 \upg{17.61}\\\textbf{82.38} \upg{17.67}\end{tabular} \\
\midrule
Food101 & \begin{tabular}[c]{@{}c@{}}11100\\7062\end{tabular} & \begin{tabular}[c]{@{}c@{}}619521\\362849\end{tabular} & \begin{tabular}[c]{@{}c@{}}56929\\31228\end{tabular} & 88.73 & \begin{tabular}[c]{@{}c@{}}\textbf{88.97} \upg{0.24}\\88.35 \downr{-0.38}\end{tabular} & \begin{tabular}[c]{@{}c@{}}88.89 \upg{0.16}\\88.87 \upg{0.14}\end{tabular} & \begin{tabular}[c]{@{}c@{}}88.87 \upg{0.14}\\88.63 \downr{-0.10}\end{tabular} & \begin{tabular}[c]{@{}c@{}}88.86 \upg{0.13}\\88.82 \upg{0.09}\end{tabular} \\
\midrule
Imagenet & \begin{tabular}[c]{@{}c@{}}198044\\21725\end{tabular} & \begin{tabular}[c]{@{}c@{}}558868\\25000\end{tabular} & \begin{tabular}[c]{@{}c@{}}502898\\61160\end{tabular} & 68.33 & \begin{tabular}[c]{@{}c@{}}69.05 \upg{0.72}\\67.04 \downr{-1.29}\end{tabular} & \begin{tabular}[c]{@{}c@{}}69.54 \upg{1.21}\\69.78 \upg{1.45}\end{tabular} & \begin{tabular}[c]{@{}c@{}}69.80 \upg{1.47}\\69.73 \upg{1.40}\end{tabular} & \begin{tabular}[c]{@{}c@{}}70.14 \upg{1.81}\\\textbf{70.33} \upg{2.00}\end{tabular} \\
\bottomrule
\end{tabular}
}
\vspace{-1mm}
\caption{Results of training with retrieval techniques $\bsetU^{\texttt{cls}}$, $\bsetU^{\texttt{cap}}$, $\bsetU^{\texttt{desc}}$ and the impact of refinement alongside the retrieved number of images.}
\label{tab:combined_data}
\vspace{-3mm}
\end{table*}

In Table \ref{tab:main_table}, we show (Top) the accuracy results of ZeroShot (ZS) CLIP and our proposed method with Linear Probing (\textit{LP}) trained on the best-retrieved dataset (from Tab. \ref{tab:combined_data}). Also, we show (Bottom) the accuracy of these methods when they are trained on the combination of the training set of each target dataset and the retrieved dataset.
We can see our method consistently outperforms ZS CLIP. We observe a significant improvement in Flowers and Cars datasets with over 15\% in accuracy, showing that the samples retrieved from the search engine best helps with the features needed to perform well on the test set of these datasets. 
When we use our method on Food and ImageNet, we obtain a slight improvement of 0.24\% and 1.81\%, respectively. These two are challenging datasets in terms of constructing queries. In the case of Food, we speculate due to the diversity and variability of the food images online and the specificity of this dataset, the improvements remained marginal. 
Regarding ImageNet, a significant challenge arises from the fact that certain classes necessitate the inclusion of highly specialized types of images. Take, for instance, classes representing specific dog species. When attempting to collect images from the web for these classes, we encounter two prevalent issues: a considerable increase in variability within the images or a prevalence of highly specific, limited-content images. 
Also, we found possible misinterpretations of the search engines in terms of concepts. Some category names could confuse search engines. For example, for the category name 
\textit{Agama} in ImageNet, which for Google could be an Indian religion or the lizard that we are looking for.
In the case of Pets, we obtain a significant improvement of more than 3\% of accuracy, surpassing the performance of training a Linear Probe on the training set of the target dataset.
\begin{figure}
\centering
\vspace{-3mm}
\includegraphics[width=0.33\textwidth]{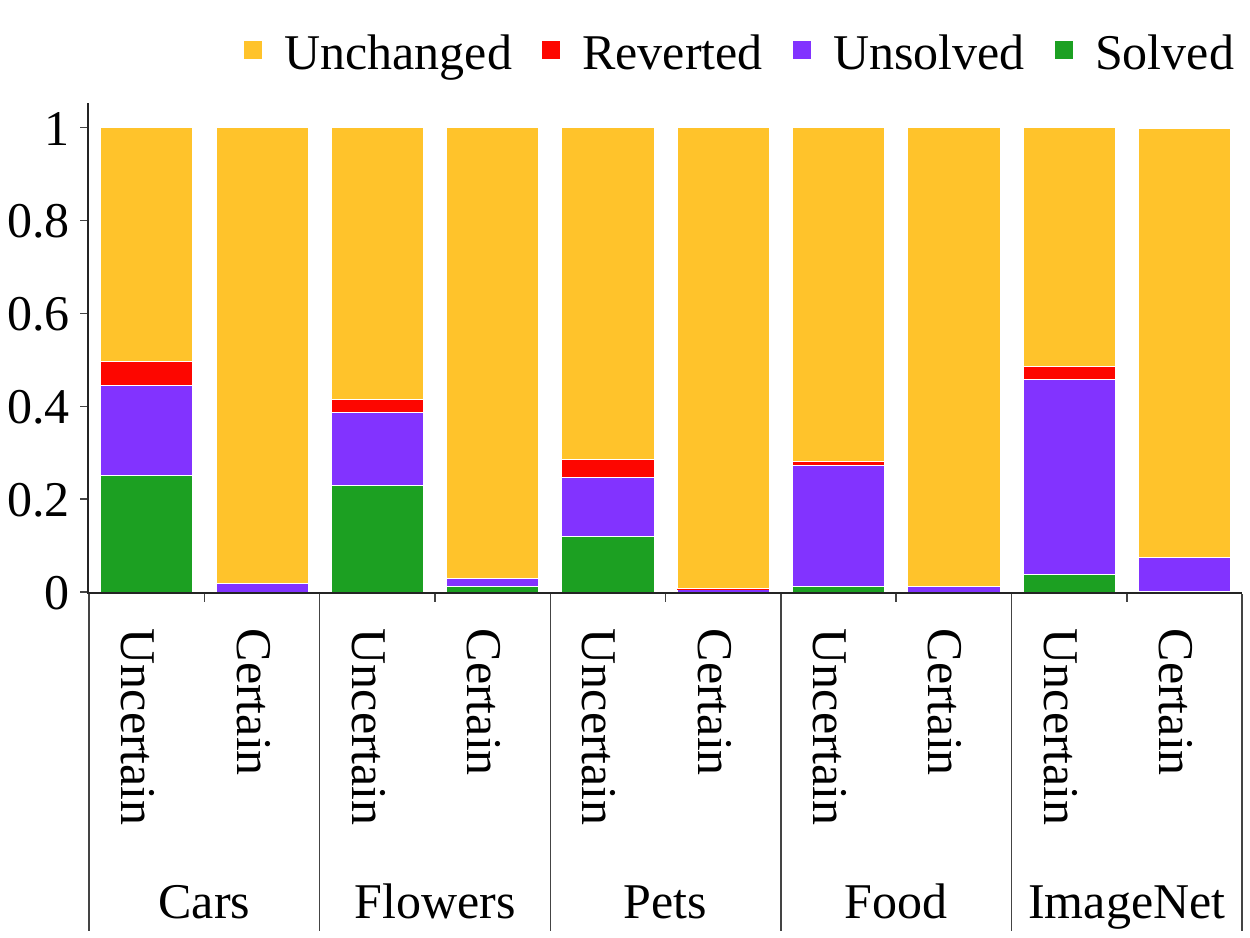}\vspace{-2mm}
\caption{\small{Analysis of improvements or certain/uncertain instances: \textbf{Solved}: shows the percentage of cases incorrectly predicted before linear probe and corrected afterwards. \textbf{Unsolved} cases are incorrectly predicted before and after when using search engine data; \textbf{Reverted}: instances predicted correctly when using CLIP but wrongly predicted after linear probe; \textbf{Unchanged}: cases were correctly predicted before fine-tuning and are still correctly predicted afterwards.}} 
\label{tab:detailed_clip_imgclass}
\vspace{-3mm}
\end{figure}

Moreover, when we use the concatenation of our retrieved dataset with the training set of the target dataset, we observe the retrieved dataset consistently improves the performance of the baseline methods when they only have access to the training set. It indicates the retrieved images provide information that supplements that of the training distribution. This suggests our approach can easily be integrated into the current pipelines.

Figure \ref{tab:detailed_clip_imgclass} shows a more detailed comparison of CLIP predictions before and after update using retrieved images. 
In this bar plot, we can see that the slight improvement in the Food dataset is due to the small margin in the reverted cases relative to the solved cases. This is while reverted cases make up a much smaller percentage of the changes in the uncertain cases of other datasets. Moreover, the effect of our method is minimal in certain cases where the percentage of unchanged cases makes up the largest fraction of every dataset. In other words, adding a Linear Probe to deal with uncertain instances does not affect the instances where CLIP is certain about the prediction.
\begin{figure*}
    \centering
    \vspace{-3mm}
    \includegraphics[width=0.65\textwidth]{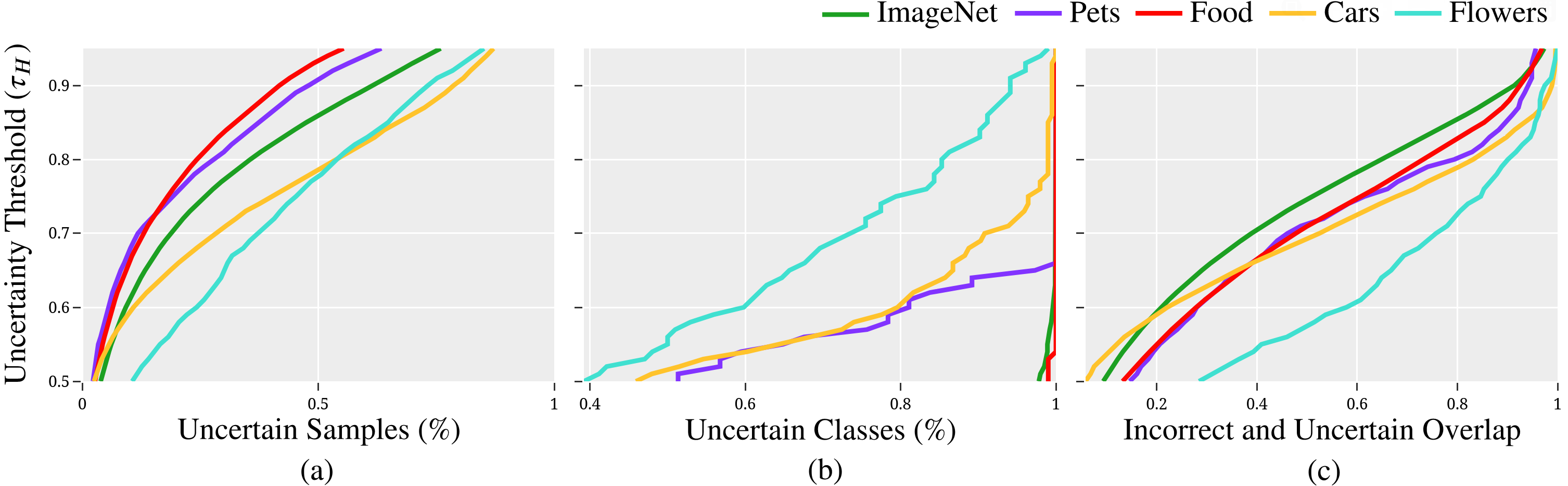}\vspace{-3mm}
    \caption{\small{Uncertain plots for each dataset: (a) percentage of uncertain instances, (b) the ratio of uncertain classes, and (c) the overlap between incorrect and uncertain instances, versus on the uncertainty threshold $\tau_H$.}}
    \label{fig:shannon_ablation}
    \vspace{-4mm}
\end{figure*}



\begin{table} 
\centering
\vspace{-1mm}
\resizebox{0.48\textwidth}{!}{
\small{\begin{tabular}{rllll}
\toprule
& \multicolumn{1}{c}{Flowers} & \multicolumn{1}{c}{Pets} & \multicolumn{1}{c}{Food} & $\times$ Images       \\ \cmidrule{2-5}
\textit{ZS CLIP}     & 66.22                       & 85.72                    & 80.97                    &              \\
IE                 & \textbf{99.10}\upg{4.50}      & 90.80\upg{5.50}          & 84.60\upg{3.60}          & $\times  10^6$ \\
Ours                            & 90.39\upg{24.17}               & \textbf{91.17}\upg{5.45} & \textbf{85.97}\upg{5.00} & $\times 10^4$    \\
\bottomrule
\end{tabular}}
}
\vspace{-2mm}
\caption{\small{Comparison with Internet Explorer (IE): Our approach excels in the Food and Pets dataset with far fewer retrieved images. Here, $\times$ Images indicates the order of magnitude difference in retrieved images. We used ResNet-50 for this comparison.}}
\label{tab:comparisonIE}
\vspace{-4mm}
\end{table} 

\subsection{On the different retrieved datasets}
Table \ref{tab:combined_data} shows the accuracy  comparison of the proposed query constructions $\bsetU^{\texttt{cls}}$, $\bsetU^{\texttt{cap}}$ and $\bsetU^{\texttt{desc}}$, Section \ref{lab:algo_steps}, and the effect of the refinement process. 
In the case of $\bsetU^{\texttt{desc}}$, we use the descriptions shared by~\cite{menon2022visual}. However, they only shared descriptions for the Flowers, Cars and ImageNet datasets.  
The effect of the refinement process does not always improve the performance of the initial retrieved dataset. Excluding Food, it always worked for datasets constructed with more diverse queries like $\bsetU^{\texttt{cap}}$. The images tend to be more diverse and cover more of the real distribution of each label when we retrieve the $\bsetU^{\texttt{cap}}$ and $\bsetU^{\texttt{desc}}$. However, those procedures also add much more noisy instances. Using refinement, we can get more samples and simultaneously remove ambiguous images that could diminish the performance of our method.

Except for the retrieved datasets $\bsetU^{\texttt{cls}}$ with refinement for Food and ImageNet, we obtain improvements compared to zero-shot CLIP-B-16 when we use the retrieved dataset. This is noticeable for the cases of Flowers and Cars, where our method obtains a large accuracy improvement on average of 14.92\% and 16.01\%, respectively.

\subsection{Comparison with Internet Explorer}
\label{subsec:comparison_with_ie}
We compare our method with Internet Explorer (IE)~\cite{internetexplorer}. In contrast with our method, which retrieved data only for the uncertain classes of CLIP and trained only with the retrieved dataset, Internet Explorer uses MoCo-V3~\cite{moco} with ResNet50 backbone pre-trained on ImageNet and also the training split of the target dataset. 
In Table \ref{tab:comparisonIE}, we compare the accuracy results of Flowers, Pets and Food since they do not report on ImageNet-1K and Cars. To fairly compare the two methods, we use CLIP with ResNet-50 backbone and train the LP with the combination of the training set of the target dataset and our retrieved dataset $\bsetU^{\texttt{cls}}$. Additionally, the improvement in accuracy is reported alongside the Top-1 accuracy values due to account for the difference in the backbones. Using two orders of magnitude of fewer data, we outperform IE on the Pets and Food dataset by 0.37\% and 1.37\%. In the case of Flower, our zeroshot baseline has a much lower performance (66.22\%) than their baseline (94.6\%). Thus, we obtain a larger margin improvement (24\%) than IE (4.5\%) on the Flower benchmark.

\subsection{On the value of the threshold $\tau_H$}
\label{sec:ablation_threshold}
\begin{figure}
    \vspace{-4mm}
    \centering
    \includegraphics[width=0.31\textwidth]{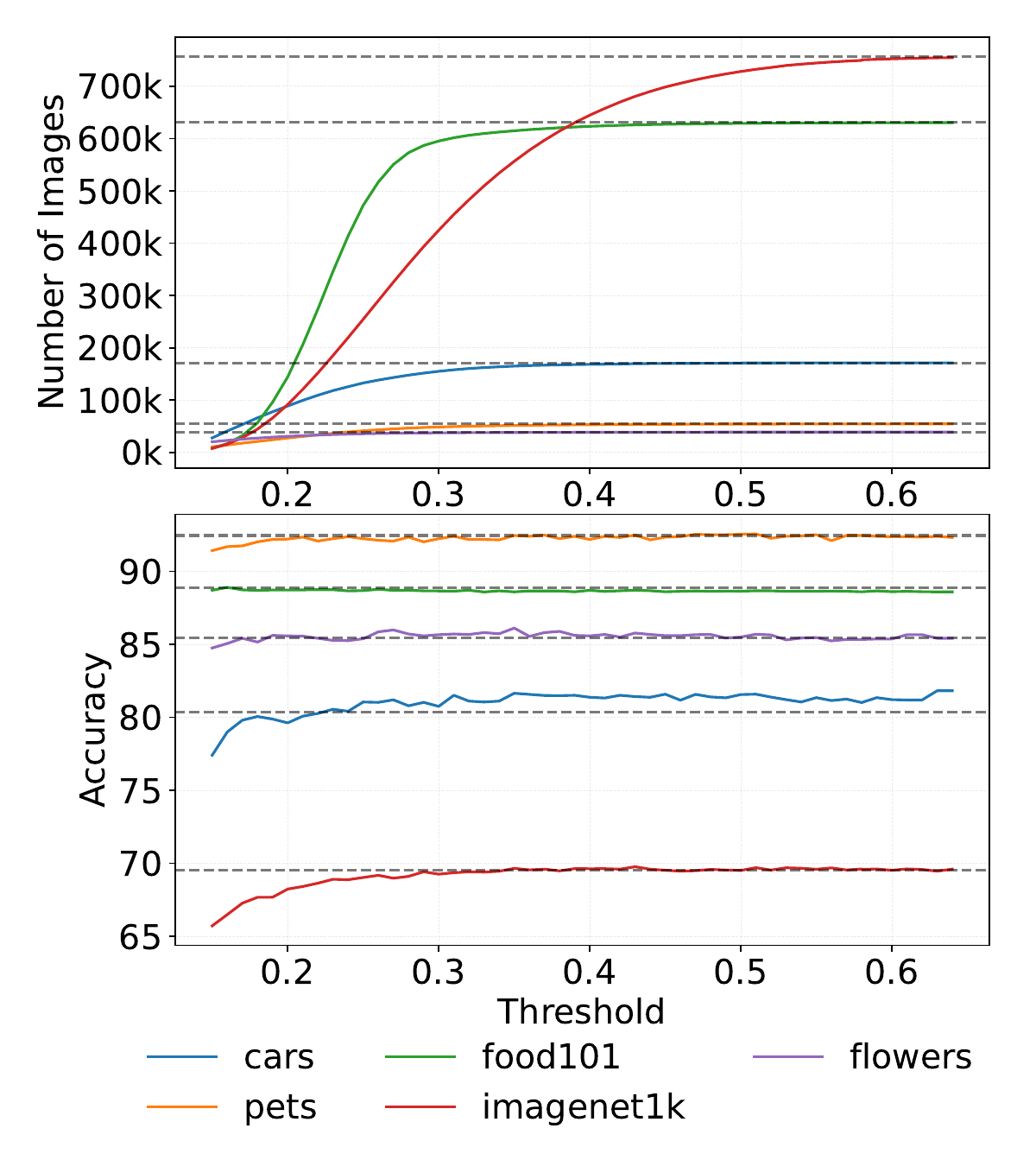}
    \vspace{-5mm}
    \caption{\small{Effect of the threshold $\tau_R$ on the number of images and accuracy. Dashed lines represent no refinement for each dataset.}} 
    \label{fig:taur_accuracy}
    \vspace{-4.5mm}
\end{figure}
To study the effects of the Shannon confidence threshold $\tau_H$ value on the uncertainty quantification process described in Section \ref{sec:uncertain}, we perform experiments on the training set of the target dataset. Figure \ref{fig:shannon_ablation} (c) shows a correlation between the incidence of incorrectly predicted samples considered uncertain for a particular $\tau_H$ and the total number of incorrectly predicted samples. This correlation highlights the utility of employing uncertainty as a surrogate measure for identifying instances of incorrect predictions. Remarkably, this relationship holds consistently across all the datasets. To maximize the capture of incorrectly predicted samples using uncertainty as a proxy, selecting a threshold value near $0.9$ would effectively achieve this objective. Moreover, Figure \ref{fig:shannon_ablation} (b) shows the percentage of uncertain classes as a factor of the threshold $\tau_H$. 
While the proportion of uncertain samples and their overlap in the Food and ImageNet datasets closely resembles that of Pets, Flowers, and Cars (as shown in Figure \ref{fig:shannon_ablation} (c)), it is noteworthy that the proportion of uncertain classes in the former two datasets nears 100\% (as depicted in Figure \ref{fig:shannon_ablation} (b)). This observation implies that, for Food101 and ImageNet, there is at least one uncertain instance for each class across a wide range of confidence thresholds. In contrast, the remaining datasets exhibit a more gradual increase in uncertainty as the confidence threshold $\tau_H$ changes, indicating that the model is more confident of some specific classes.



\subsection{On the refinement process}
\label{subsec:umap}
\textbf{The value of threshold $\tau_R$}. Figure \ref{fig:taur_accuracy} shows the effect of a strict to a more relaxed refinement threshold $\tau_R$ on the accuracy of our refinement process in the retrieved dataset $\bsetU^{\texttt{cap}}$. Notice that small values in $\tau_R$ mean we reduce the acceptance range, Section \ref{sec:refinement}. We use a threshold of $0.45$ across all the datasets, where we observe performance improvement in a variety of benchmarks whilst not removing a big portion of the retrieved dataset, thus covering the target dataset distribution, see Appendix \ref{app:umap}.

\begin{figure}
\vspace{-3mm}
\centering
\includegraphics[width=0.4\textwidth]{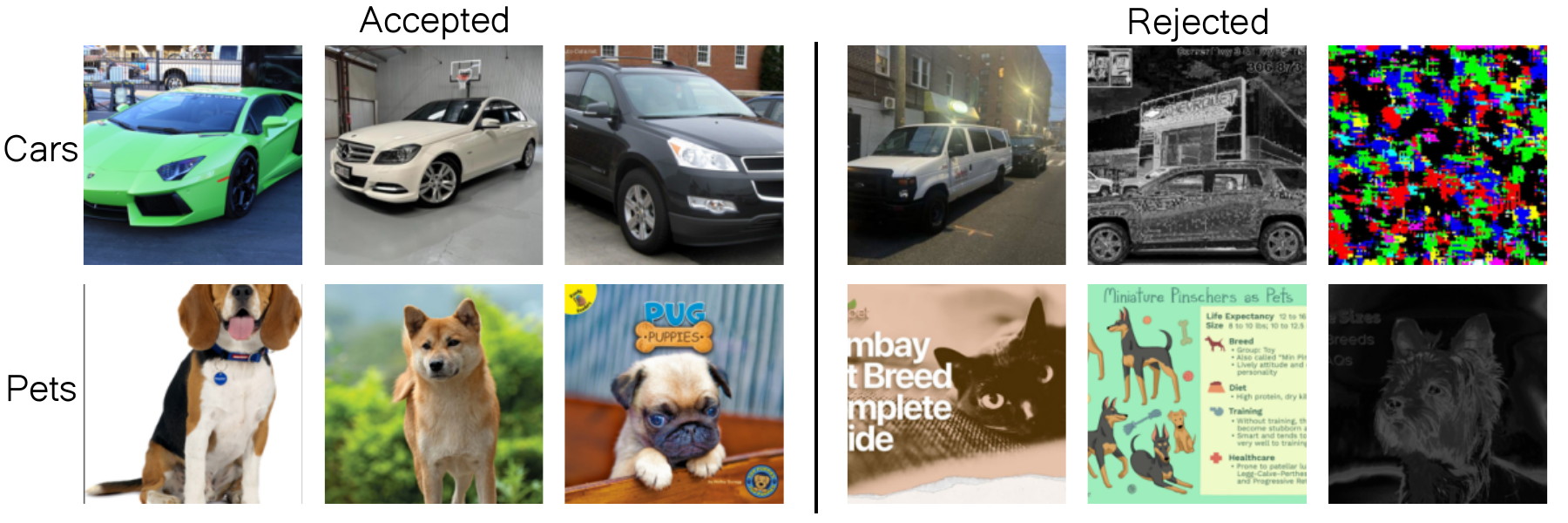}
\vspace{-3mm}
\caption{\small{Accepted/rejected examples using our refinement.}}
\label{fig:qualitative_refinement}
 \vspace{-2mm}
\end{figure}

\textbf{Qualitative analysis of refinement} Figure \ref{fig:qualitative_refinement} shows random samples of our refinement process's accepted and rejected instances for Pets and Food datasets. 
Other datasets are in the Appendix \ref{app:qualitative_analysis_refinement}. 
We can see that the refinement process can reject images that contain words related to the class names on the dataset but do not necessarily have the object in interest, thus not providing information to our classifier. It is also capable of removing instances that contain the object, like Siamese cats, but they are out of the distribution of the training dataset.


\subsection{On using Parameter-Efficient Fine-Tuning}
\label{app:lora}

\begin{table}
\centering
\resizebox{0.5\textwidth}{!}{%
\begin{tabular}{rllll}
\toprule
\multicolumn{1}{r}{}                                                                            & \multicolumn{1}{c}{Flowers} & \multicolumn{1}{c}{Pets} & \multicolumn{1}{c}{Cars} & \multicolumn{1}{c}{Food} \\ 
\cmidrule(lr){2-5}
\textit{ZS CLIP-B-16}                                                                  & 71.15                       & 89.04                    & 64.71                    & 88.73                  \\ 
\cmidrule(lr){1-5}
 \textit{Using LAION-5B}& 61.86 \downr{-9.29}         & 80.84 \downr{-8.20}      & 56.60 \downr{-8.10}      & 88.35 \downr{-0.38}                                    \\
\cmidrule(lr){1-5}
\textit{LP w/ $\bsetU^{\texttt{cls}}$}                                            & 86.24 \upg{15.09}           & 92.61 \upg{3.57}         & 80.52 \upg{15.81}        & 88.97 \upg{0.24}      \\

\textit{\begin{tabular}[c]{@{}c@{}}LoRa w/ $\bsetU^{\texttt{cls}}$\end{tabular}}                 & 84.29 \upg{13.14}           & 92.56 \upg{3.52}         & 80.55 \upg{15.84}        & 88.98 \upg{0.25}     \\
\bottomrule
\end{tabular}
}
\vspace{-1mm}
\caption{\small{Accuracy results of training a LP vs using \textbf{LoRa} on $\bsetU^{\texttt{cls}}$ and also the results of using LAION-5B to retrieved images and train our LP.}}
\label{tab:lora}
\vspace{-5mm}

\end{table}
LoRA~\cite{lora} is a prominent parameter-efficient fine-Tuning method primarily employed in transformers to avoid the need for retraining the entire model for downstream tasks. Instead of retraining the entire model from scratch for each task, LoRA preserves the pre-trained model and introduces smaller, trainable matrices to each layer of the model. These matrices enable the model to adjust to various tasks without modifications to all of the parameters. We use LoRa to adapt CLIP with our retrieved dataset $\bsetU^{\texttt{cls}}$ as an alternative to LP. Table \ref{tab:lora} shows the performance of CLIP-LoRa. As observed, LP outperforms CLIP-LoRa in both Flowers and Pets benchmarks. In the cases of Cars and Food, the performance gain is marginal. Remarkably, using the retrieved dataset $\bsetU^{\texttt{cls}}$ always helps LoRa to adapt. This indicates more parameters of LORA compared to LP overfits to the retrieved dataset.

\subsection{LAION-5B as an alternative to search engines}
\label{app:searching_image}
Instead of using the language to create queries and search images, we propose an alternative approach that uses the LAION-5B~\cite{schuhmann2022laion} to retrieve similar images in the multimodal space. Since we cannot access the training set to get the labels of the most uncertain cases, we retrieved images from the most certain image for CLIP. In this case, we aim to retrieve the neighbour images similar to those of certain instances so that we can assign the label of the certain sample to the retrieved one. As shown in Table \ref{tab:lora}, naively retrieving similar images and training a linear probe does not help improve predictions on the target dataset. 

\section{Evaluating possible data leaks}
\label{app:eval_data_leaks}
A major apprehension when using search engines to build a dataset revolves around performance enhancement being predominantly driven by retrieving a significant portion of the test set images. We mitigate these concerns by assessing the extent to which the retrieved images match those found in the test set of the target dataset using Difference Hashing (dHash) \cite{buchner_imagehash}---a hash comparison method. 
Our method retrieves a minuscule portion of test set images online, thus no leakages. We also report the results of LP on all uncertain datasets after removing every leaked image to show our method's substantial consistent improvement cannot be attributed to test set leakage.

\begin{table}
\vspace{-2mm}
\centering
\resizebox{\columnwidth}{!}{%
\begin{tabular}{cccccc}
\toprule
                         & Flowers                     &  Pets                    & Cars                        & Food                     & ImageNet \\ \midrule
Target test set size     & 6148                        &  3669                    & 8041                        &  25250                   & 50000    \\
$\bsetU^{\texttt{cls}}$  & 5 {\color{blue}($\approx$ 0.0\%)} &  2 {\color{blue}($\approx$ 0.0\%)} & 105 {\color{blue}(1.3\%)} &  0 {\color{blue}(0.0\%)} & 107 {\color{blue}($\approx$ 0.0\%)}  \\
$\bsetU^{\texttt{cap}}$  & 4 {\color{blue}($\approx$ 0.0\%)} &  7 {\color{blue}($\approx$ 0.0\%)} & 448 {\color{blue}(5.5\%)} &  0 {\color{blue}(0.0\%)}   &  121 {\color{blue}($\approx$ 0.0\%)}      \\
$\bsetU^{\texttt{desc}}$ & 5 {\color{blue}($\approx$ 0.0\%)} & 17 {\color{blue}($\approx$ 0.0\%)} & 245 {\color{blue}(3.0\%)} &  0 {\color{blue}(0.0\%)}   & 184 {\color{blue}($\approx$ 0.0\%)}   \\
Accuracy Drop & 0.01 & 0.00 & 0.08 &  \xfill{.1em} & 0.00 \\
\bottomrule
\end{tabular}%
}
\caption{Number of leaked images from the test set with the percentage of the test set in the leaked images in blue.}
\label{tab:dataleak}
\vspace{-2em}
\end{table}

\section{Conclusion}
The proposed approach of leveraging search engines for machine learning is a novel idea with connections to different areas of research, including active learning, weakly supervised learning, cross-modal retrieval, and language models. In the future, we envision a dynamic approach where pre-trained models can access the internet to acquire new knowledge as they encounter novel concepts for prediction. This stands in contrast to traditional task-specific static models, which are expected to perform adequately even in the presence of previously unseen instances, a challenge that current out-of-distribution generalization research strives to address. 

Much like the art of effective prompt engineering, our methodology suggests that formulating well-constructed search queries (attainable \eg through interaction with LLMs) may yield improved performance, warranting further exploration. It is worth noting that our approach may encounter limitations in scenarios with restricted online resources, such as specialized medical domains, where purpose-built search engines could be advantageous.

One avenue for potential future research involves amalgamating results from various search engines. This could lead to divergent sets of retrieved data. Additionally, the susceptibility to bias is a pertinent concern with such datasets. As these methods mature, addressing such issues becomes increasingly important.
{\small
\bibliographystyle{ieee_fullname}
\bibliography{main}
}

\clearpage
\appendix

\section{Visualization of the target versus retrieved dataset}
\label{app:umap}
\begin{figure}[h]
    \centering
        \includegraphics[width=0.3\textwidth]{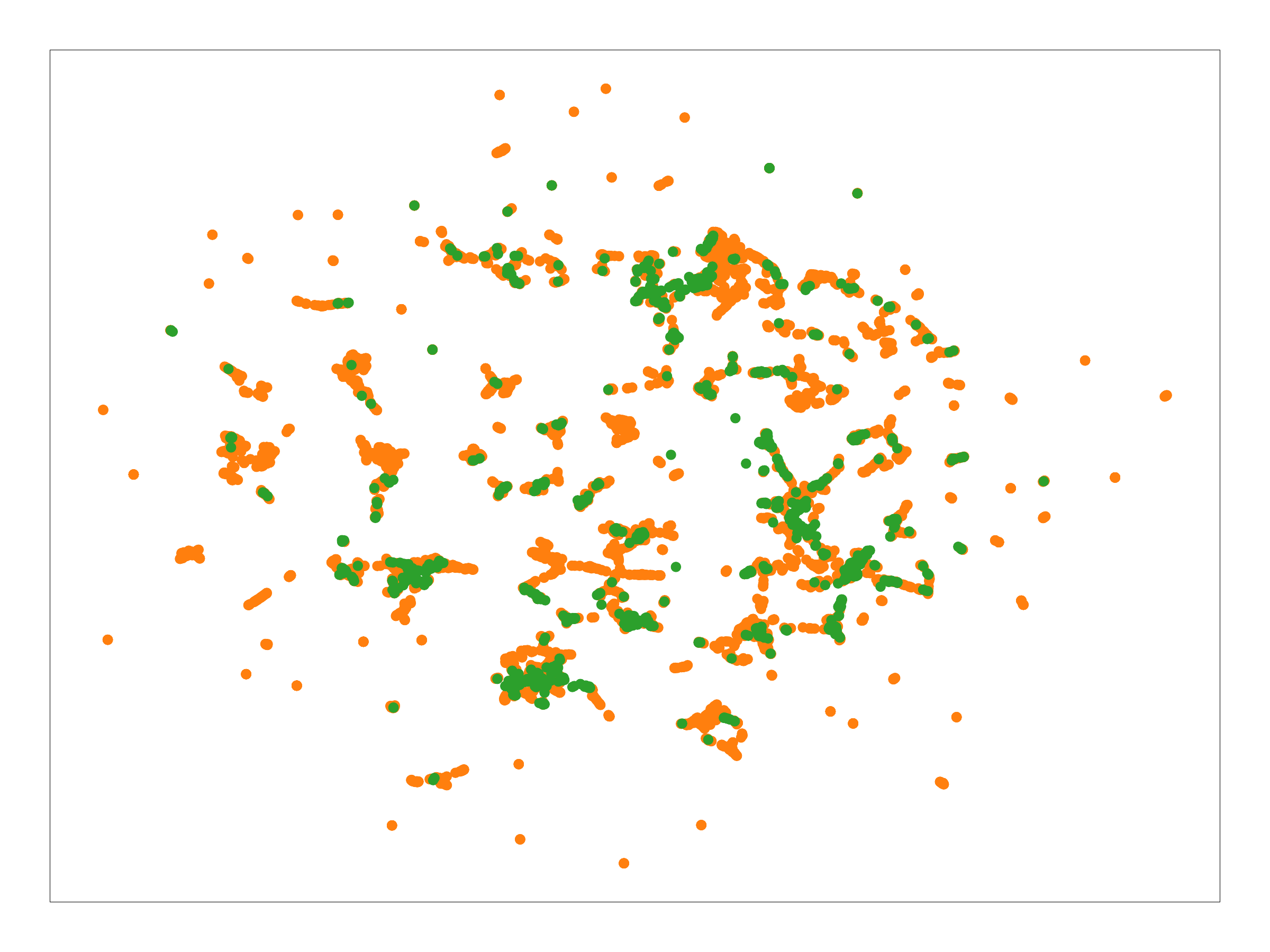}
        \includegraphics[width=0.3\textwidth]{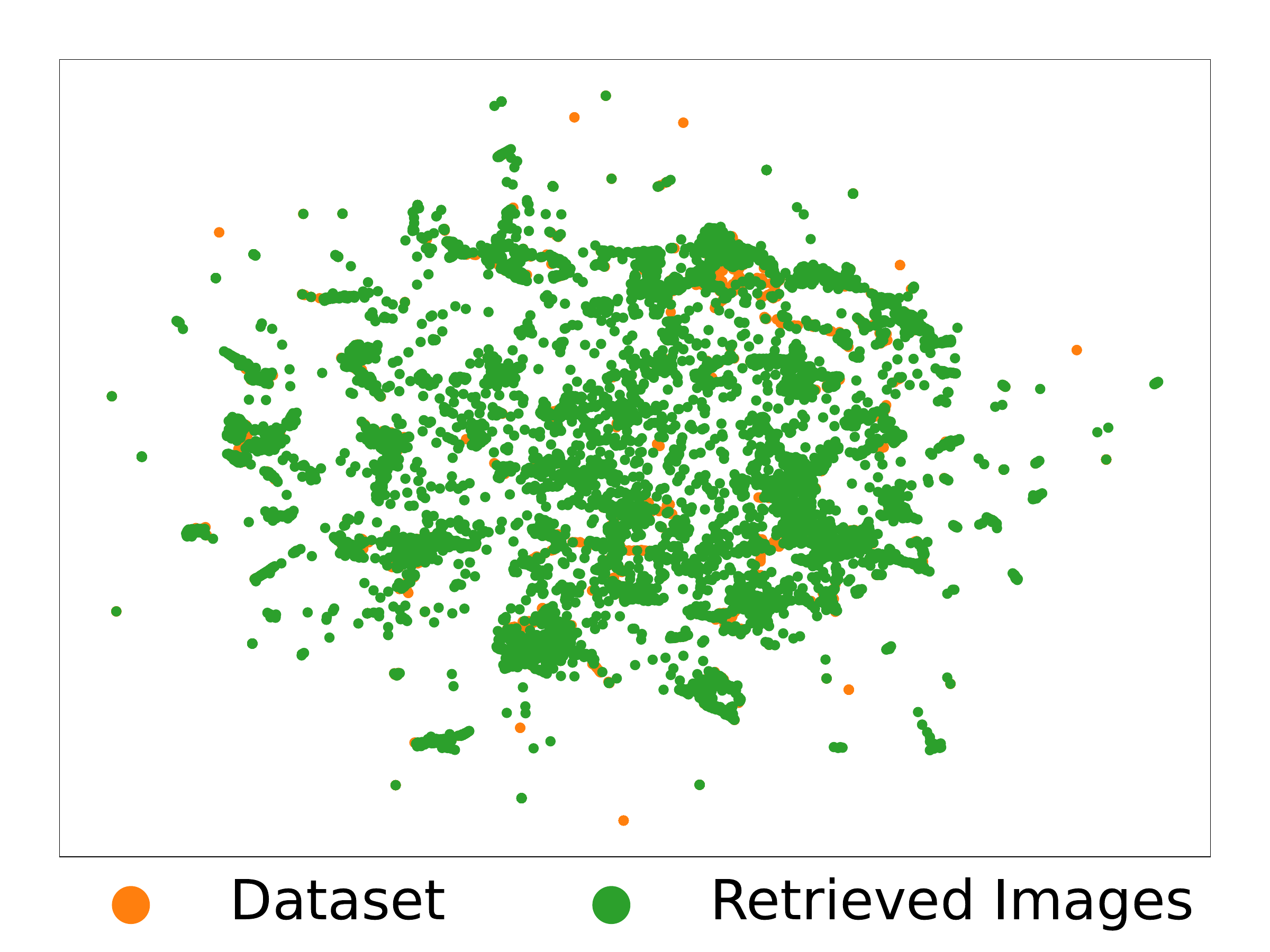}
    \caption{UMAP visualization of the distribution of the image features from CLIP-B-16 for the training split of the original and retrieved dataset. It shows the effect of the choice of the threshold $\tau_R$ refinement method on the retrieved dataset from a less (bottom) to a more strict (top) refinement.} 
    \label{fig:umap_distribution}
\end{figure}

Figure \ref{fig:umap_distribution} shows the UMAP visualization of the image features of the target dataset and the retrieved dataset for two choices of the threshold $\tau_R$. First, we can see that the retrieved dataset matches the distribution of the training set. The retrieved dataset can fill the gap in the distribution of each class. This explains why our LinearProbe can improve the performance of CLIP in uncertain cases. Moreover, we can see how a more strict threshold $\tau_R$ retained less number of instances in the retrieved set. A balanced compromise can be achieved by selecting a retrieved dataset that best overlaps the target distribution while excluding noisy samples.



\section{Effects of prompt specificity on prediction confidence}
\label{sec:prompt_specificity}
\begin{figure}[h]
    \centering
    \begin{subfigure}[b]{.49\textwidth}
        \includegraphics[width=0.99\textwidth]{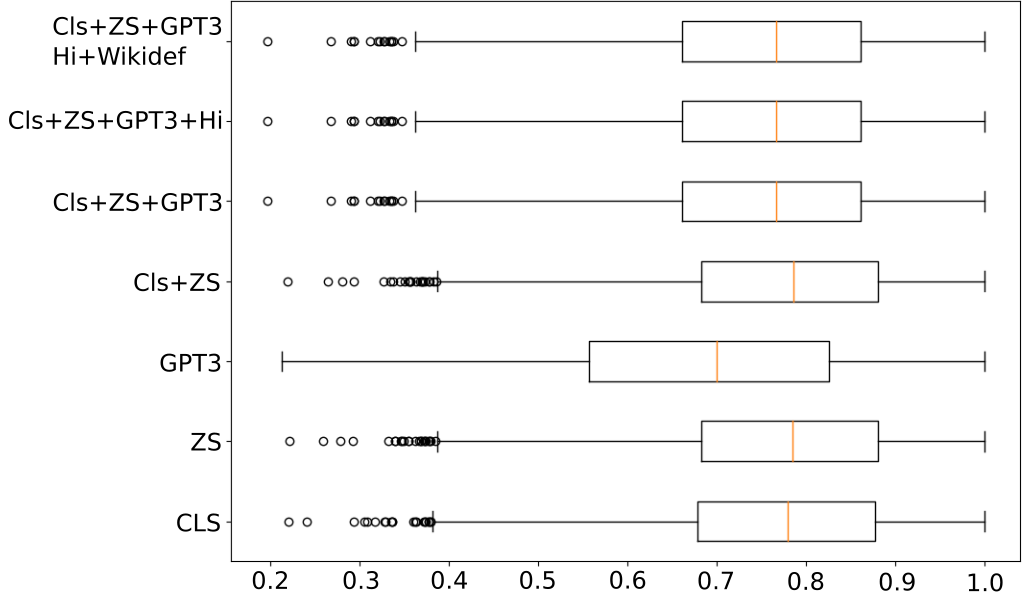}
        \caption{Cars dataset}
    \end{subfigure}
    \begin{subfigure}[b]{.49\textwidth}
        \includegraphics[width=0.99\textwidth]{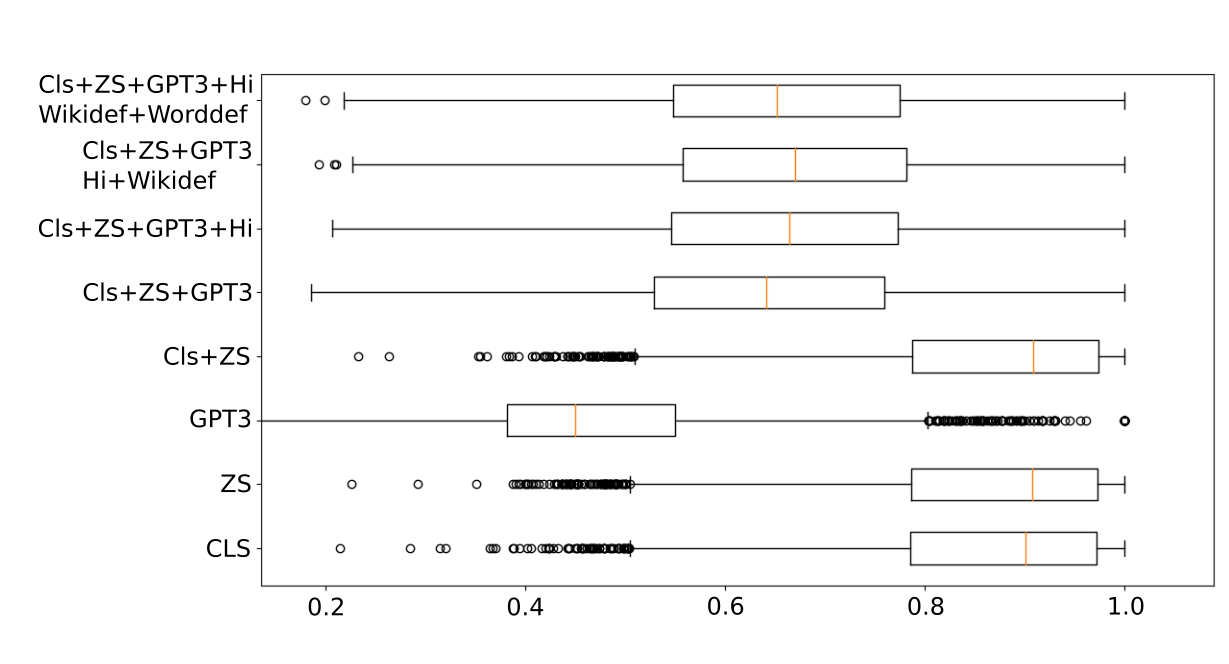}
        \caption{Pets dataset}
    \end{subfigure}
    \caption{Boxplot of confidence values when prompts with varying specificity are used. \textbf{CLS} refers to using class names from the dataset for prompting. \textbf{ZS} are prompts specific to each dataset described in more detail in Appendix \ref{sec:prompt_specificity}. \textbf{GPT3}, \textbf{Wikidef}, \textbf{Hi} and \textbf{Worddef} are prompt templates taken from \cite{li2022elevater}. \textbf{GPT3} refers to descriptions of each class generated by OpenAI-GPT3 \cite{pretrain-zeroshot} and used as the prompt. \textbf{Worddef} and \textbf{Wikidef} are Wordnet \cite{wordnet} and Wiktionary \cite{wiktionary} definitions of each class in the dataset (With the exception of Cars for which no classes have Wordnet definitions). \textbf{Hi} refers to a set of words extracted from Wordnet based on the ontological hierarchy of the class name. } 
    \label{fig:confidence_boxplot}
\end{figure}

Since predictions from CLIP are reliant not only on the input images but also on the text prompt, the uncertainty of the prediction will also be affected by this text. To study the effects of the specificity of the input prompt on the confidence in the prediction, we use six methods of prompting, each at different levels of descriptiveness. The results of this study are visualized in Fig. \ref{fig:confidence_boxplot}.

The prompt named CLS refers to using class names with the format ``\texttt{\{class\_name\}}" to prompt CLIP. ZS expands this approach by adding a generic description of the dataset in the format ``\texttt{\{class\_name\}, a type of \{description\}}", e.g. the prompt used for Pets is ``\texttt{\{class\_name\}, a type of pet}". For GPT3, a description of the class name is generated by OpenAI GPT3 first, and the description is used for prompting without the addition of the class name that was used to generate the description or any other phrases. Wordnet and Wikidef refer to the definition of the class name in WordNet~\cite{wordnet} and Wiktionary~\cite{wiktionary}. Similar to GPT3, no additional phrases were added for prompting. Finally, Hi refers to generating a prompt using the WordNet hierarchy where the last three words in the ontological tree leading to the class name are concatenated and used as the prompt.

Fig. \ref{fig:confidence_boxplot} indicates that utilizing ZS individually, or in combination with CLS, marginally enhances the network’s confidence in predictions. This enhancement is attributable to a more refined representation of the prompts, thereby reducing text ambiguity. A notable observation is the significant reduction in network confidence with the use of GPT-3 descriptions compared to simpler prompts such as CLS or ZS. An examination of these descriptions revealed that GPT-3 prompts tend to be more narrative and detailed rather than focusing on keyword specificity. That is, instead of representing a concise combination of the class’s crucial characteristics, these prompts offer extensive narratives, often overlooking the visually distinguishable aspects of the class.

\section{On the number of images per query}
\label{sec:num_image_ablation}

\begin{figure}[h]
    \hspace{-7mm}
    \includegraphics[width=0.565\textwidth]{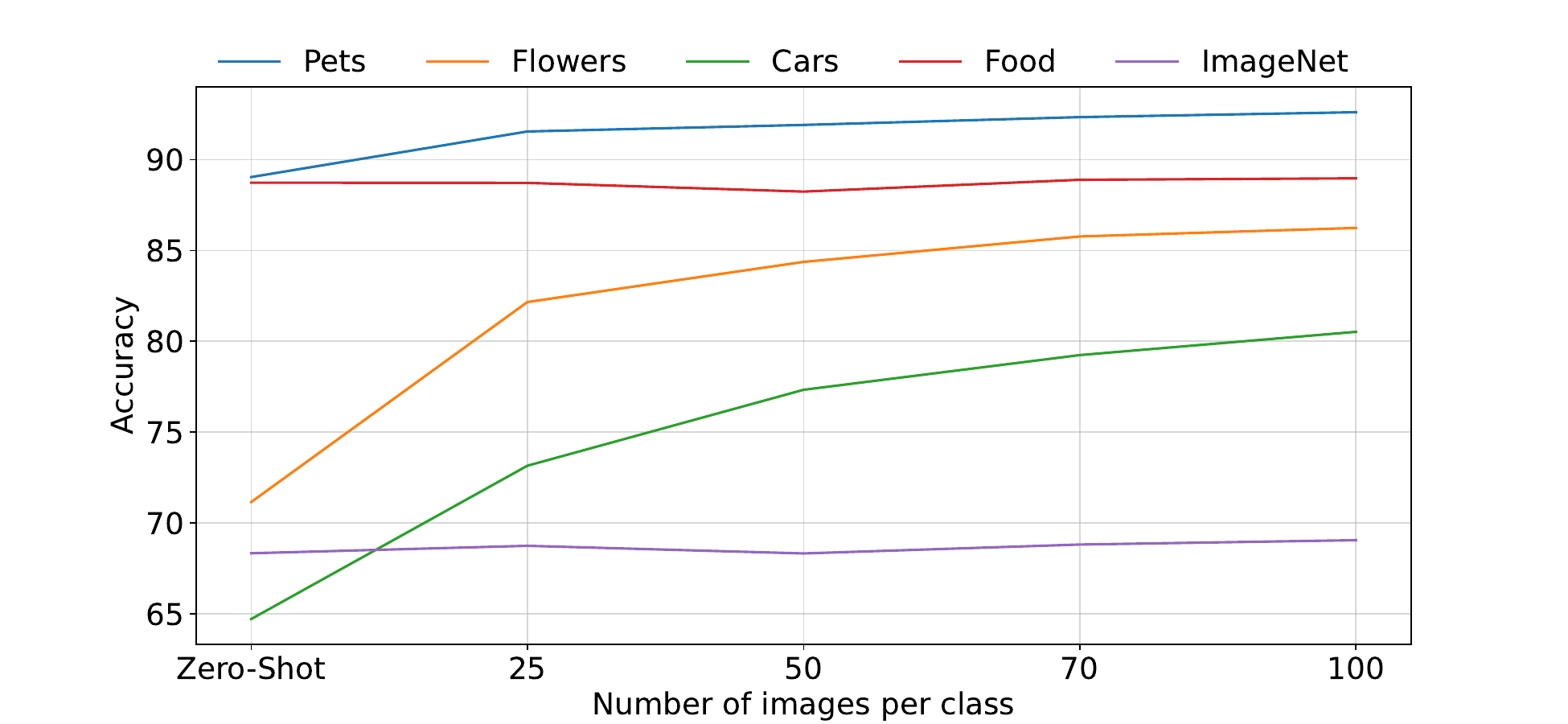}
    \caption{Exploring the impact of varying the number of images utilized in training a LinearProbe.}
    \label{fig:num_images}
\end{figure}

To investigate the effect of the number of images retrieved per class, we train a classifier on subsets of increasing size from $\bDyU^{cls}$ and report the accuracy results for each dataset in Fig. \ref{fig:num_images}. For Cars, Flowers and Pets we observe a steady increase in accuracy as the number of images from the retrieved set used for training increases. This is while the lowest number of used images (25 for all datasets) still outperforms the zero-shot accuracy for every dataset except Food. In ImageNet, on the other hand, we observe a slight decline in accuracy when using 50 images, but the overall trend is upward with more images.

Furthermore, in Cars, Flowers, and Pets datasets, the rate of accuracy improvement in relation to the increase in the number of images diminishes as more data is incorporated, eventually reaching a plateau at around 100 images. This phenomenon arises from insufficient additional information in newer images or the limitations of the search engine in retrieving more relevant images for the classifier.


\section{Other refinement methods}
The refinement method described in Section \ref{sec:refinement} shows the exclusion of unrelated and noisy images from the retrieved set; however, this method is intrinsically dependent on the training split of the target dataset. In this experiment, we investigate an alternative refinement approach that is not dependent on accessing images from the dataset and explore an approach that circumvents the computationally demanding clustering strategy depicted in Section \ref{sec:refinement}.
\begin{figure}
    \centering
    \begin{subfigure}[b]{.5\textwidth}
        \includegraphics[width=0.99\textwidth]{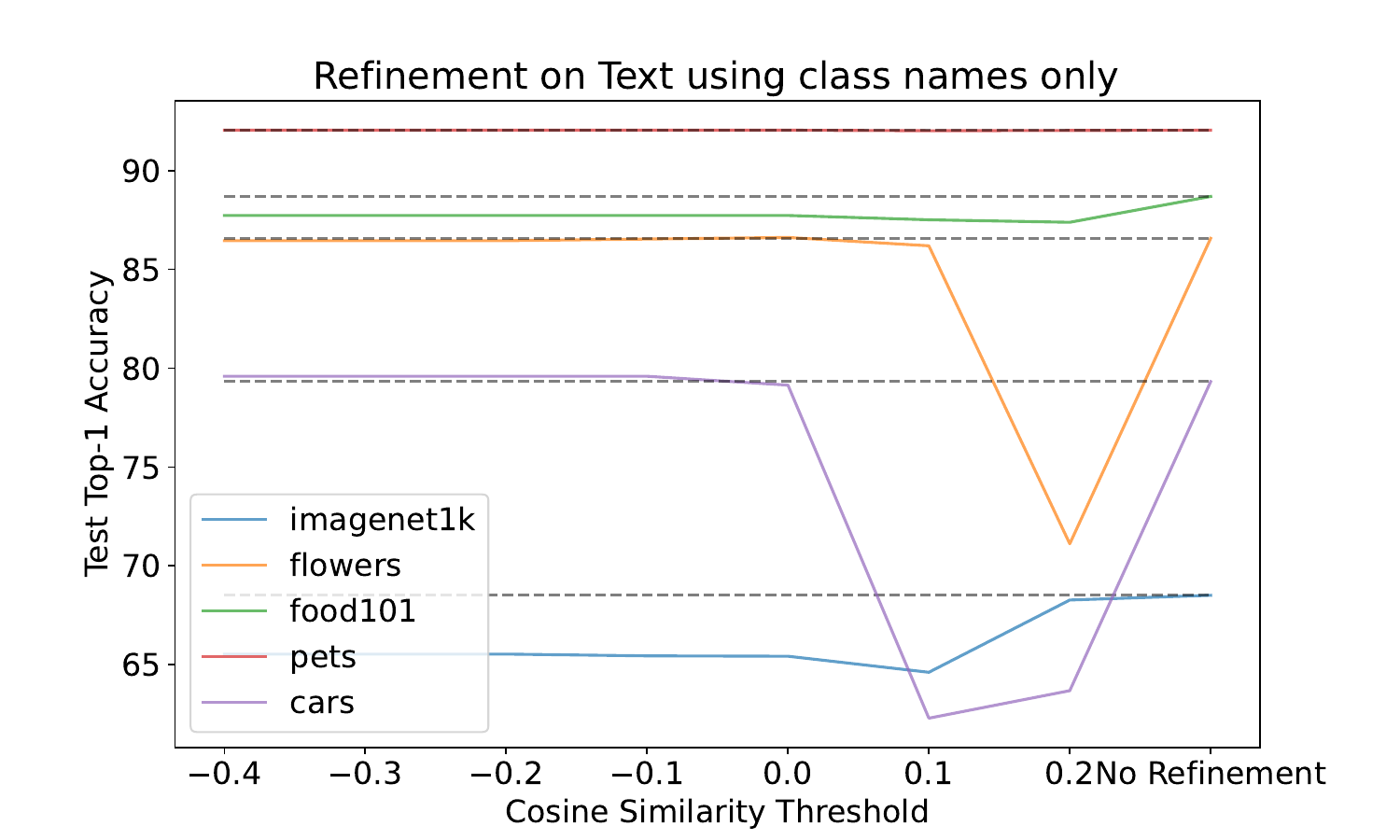}
        \caption{Refinement on $\bsetU^{\texttt{cls}}$}
    \end{subfigure} \vfill
    \begin{subfigure}[b]{.5\textwidth}
        \includegraphics[width=0.99\textwidth]{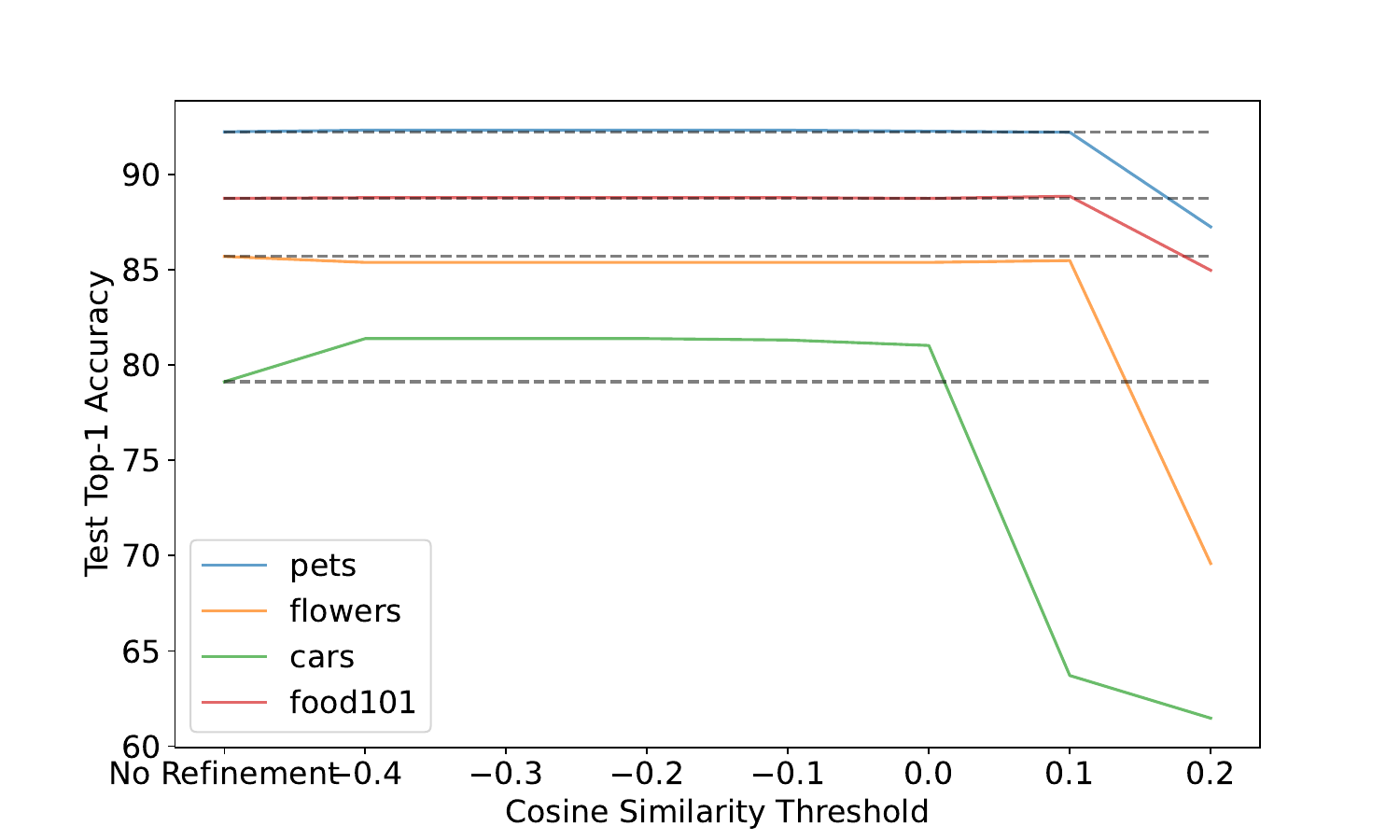}
        \caption{Refinement on $\bsetU^{\texttt{cls+cap}}$}
    \end{subfigure}
    \caption{Refinement using the labels of the target dataset} 
    \label{fig:text_refinement}
\end{figure}
\begin{figure}
    \centering
    \begin{subfigure}[b]{.5\textwidth}
        \includegraphics[width=0.95\textwidth]{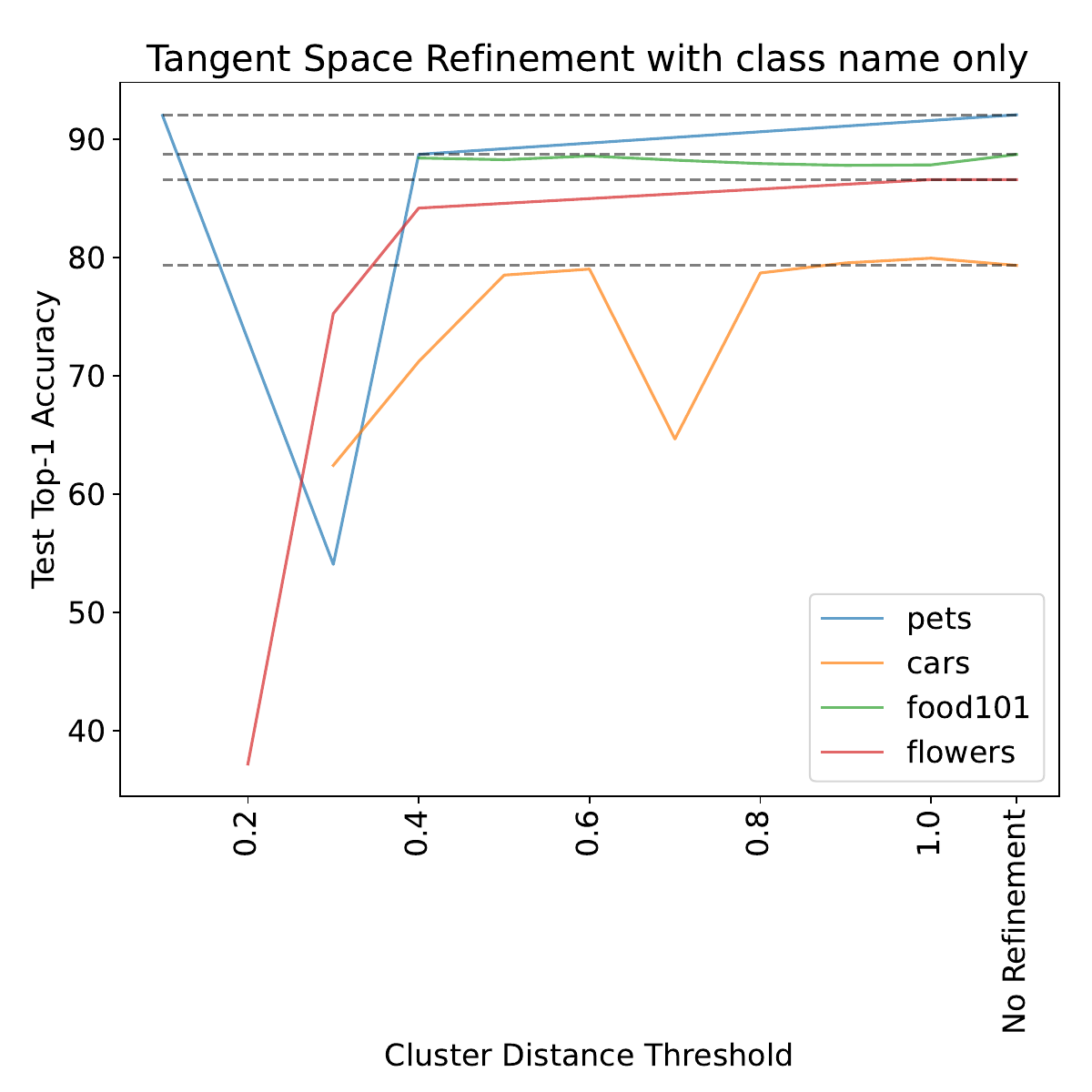}
        \caption{Refinement on $\bsetU^{\texttt{cls}}$}
        \label{fig:tangent_refinement_top}
    \end{subfigure} \vfill
    \begin{subfigure}[b]{.5\textwidth}
        \includegraphics[width=0.95\textwidth]{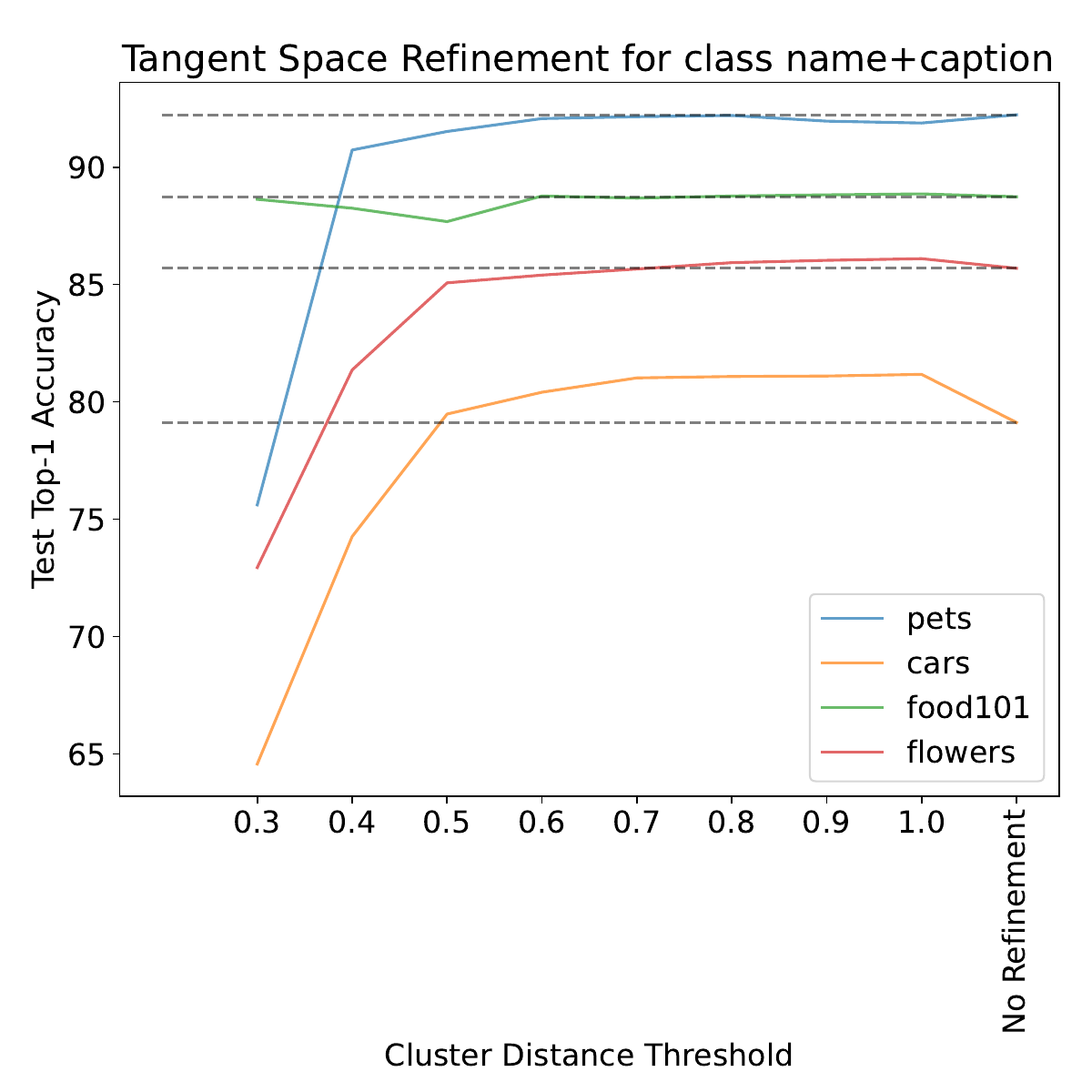}
        \caption{Refinement on $\bsetU^{\texttt{cls+cap}}$}
        \label{fig:tangent_refinement_bottom}
    \end{subfigure}
    \caption{Refinement using the tangent space of the image modality with k-means clustering} 
    \label{fig:tangent_refinement}
\end{figure}

For refining retrieved images, we need anchor points that facilitate gauging the similarity of each retrieved image to the target data distribution. Even without access to images from the target dataset, textual values of labels for the target dataset can still serve as a viable alternative. Essentially, this process involves conducting inference on the retrieved images, identifying text with the highest similarity to each image, and discarding those images with their similarity value falls below a predefined threshold, denoted as $\tau_R$ according to the following equation:
{\begin{eqnarray}
f(\bx_r\mid\by) & = & \mathbb{I}\big[\langle \phi_x(\bx_r), \phi_t(\by_{i^*}) \rangle  < \tau_R\big], \\ 
i^* &=& \arg\max_i\, \langle \phi_x(\bx_r), \phi_t(\by_{i^*}) \rangle.    
\end{eqnarray}}
On the other hand, as a substitute for the clustering on the unit hypersphere methodology detailed in Section \ref{sec:refinement}, we investigate the application of the more efficient k-means clustering. Recognizing that k-means is not inherently designed for clustering on a sphere, we start by elevating each point on the hypersphere to the tangent space at the mean of the image modality. This mean is obtained using the images from the training split of the target dataset. 
Consequently, the overall procedure for this strategy is similar to the one described in Section \ref{sec:refinement} with the exception of employing the Euclidean tangent space, thereby facilitating clustering using k-means. The outcome of the text-based approach is visualized in Fig. \ref{fig:text_refinement} and for the k-means approach in Fig. \ref{fig:tangent_refinement}.

While the approach proposed in Section \ref{sec:refinement} gives better results than the ones seen in Fig. \ref{fig:text_refinement} and \ref{fig:tangent_refinement}, we still see some improvement especially for Cars in both figures. Two observed issues with the tangent space clustering are the issues with the distortion in the projection of images to the tangent space and the other is the problems with failure of k-means in finding appropriate cluster centers leading to large decline in the accuracy observed in Fig. \ref{fig:tangent_refinement_top}. 


\section{Image distributions comparison}
\label{sec:dist_vis}

\begin{figure}
    \centering
    \begin{subfigure}{0.4\textwidth}
        \centering
        \includegraphics[width=1\linewidth]{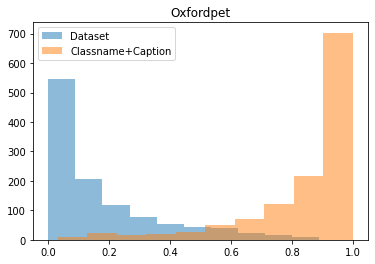}
        \caption{Pets image distribution}
        \label{fig:sub1}
    \end{subfigure}
    \hfill
    \begin{subfigure}{0.4\textwidth}
        \centering
        \includegraphics[width=1\linewidth]{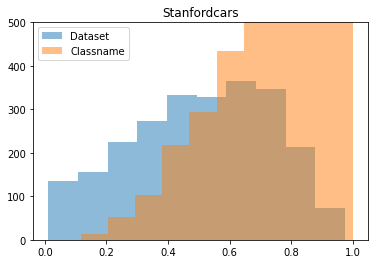}
        \caption{Cars image distribution}
        \label{fig:sub2}
    \end{subfigure}
    \vspace{1em} 
    \begin{subfigure}{0.4\textwidth}
        \centering
        \includegraphics[width=1\linewidth]{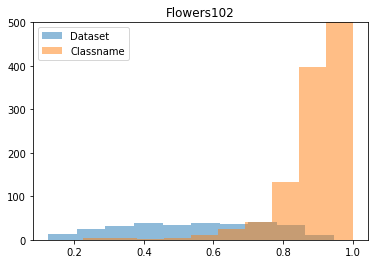}
        \caption{Flowers image distribution}
        \label{fig:sub3}
    \end{subfigure}
    \hfill
    \begin{subfigure}{0.4\textwidth}
        \centering
        \includegraphics[width=1\linewidth]{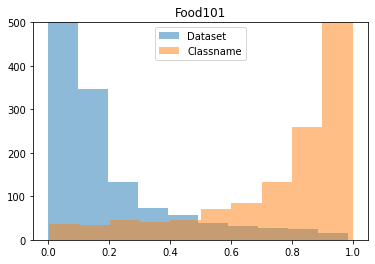}
        \caption{Food image distribution}
        \label{fig:sub4}
    \end{subfigure}
    \caption{Distribution of the retrieved images compared to the images from the target dataset.}
    \label{fig:dist_vis}
\end{figure}

To qualitatively investigate the differences between the distribution of images from the target dataset to that of the retrieved dataset, we train a classifier to detect which dataset the images are coming from. In particular, analogous to a discriminator in a generative adversarial network (GAN), we train a binary classifier to distinguish the retrieved dataset from the target one.
The outputs of this classifier are shown in Fig. \ref{fig:dist_vis} for different datasets. We anticipate that instances characterized by increased overlap--indicative of greater confusion by the classifier--will exhibit a more closely aligned distribution. For the Food and Pets dataset, there is minimal overlap, suggesting that the retrieved data exhibits less similarity.

We hypothesize that this easy distinction by the classifier is because of a combination of reasons, one of which is a genuine shift between the distribution of retrieved and target images. This could be due to the lack of specificity in the class names used to retrieve images. Consider the scenario of retrieving images of cars produced by a specific manufacturer. In this case, images of the same cars, sharing the same make and model, from both the retrieved and target datasets may exhibit little visual distinction. This is attributed to their common manufacturer and the descriptive nature of the class names, which facilitates a close match between the images in the two datasets.
On the other hand, when it comes to the Food or Pets dataset, images retrieved by a search engine will be much more diverse than their corresponding curated ones, which indicates biased distributions. Therefore, distribution shifts are expected for these datasets. The Flowers dataset seems to have a good overlap while retrieving many distinct images. 

Another potential reason for this distinction could be much shallower than the previously suggested one. The images from the target dataset are curated images that commonly have a fixed height-to-width ratio. On the other hand, retrieved images will have different ratios even between themselves. While resizing and cropping images before inputting them into CLIP, there's a potential for consistent content cropping in the retrieved images, whereas the entirety of the object's content in the target dataset remains intact. This could serve as a superficial cue for the classifier, potentially leading to shortcut learning.

\section{Quantitative measure of difference between retrieved and target datasets}
\begin{figure}[h]
    \centering
    \includegraphics[width=0.475\textwidth]{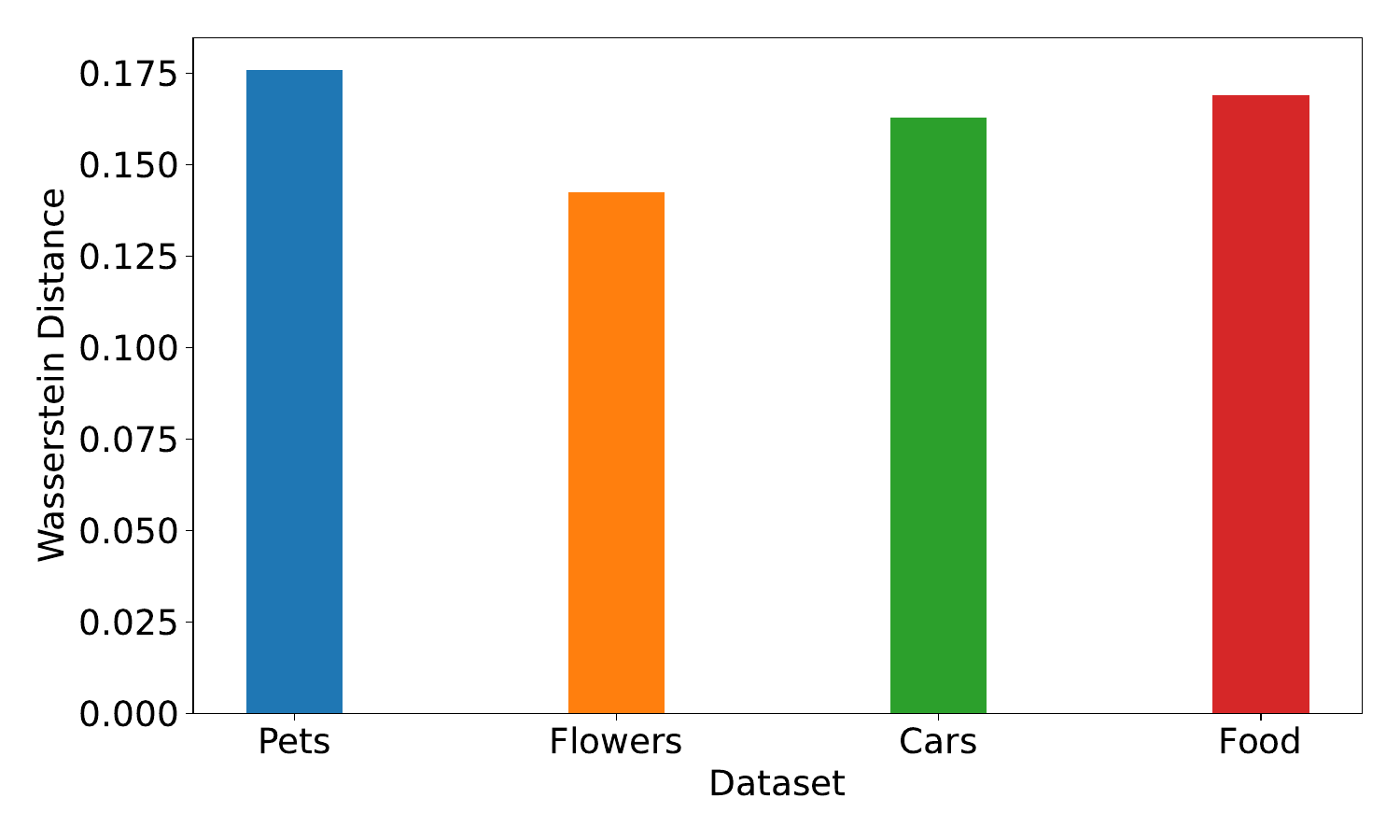}
    \caption{EMD between the feature distribution of the retrieved and target images}
    \label{fig:wasserstein}
\end{figure}
To quantify the difference between the distribution of features of the retrieved images compared to that of the target dataset, we use the Earth Mover's Distance (EMD)~\cite{Rubner2000TheEM}. EMD or the Wasserstein distance between the two distributions can show how different the retrieved images are in the feature space compared to the target dataset by quantifying the minimum cost of turning one distribution into the other, where cost is measured in terms of moving mass through the space. The value for Pets, Cars, Flowers and Food datasets is visualized in Fig. \ref{fig:wasserstein}.
Comparing the datasets to each other, Flowers has the lowest difference between the retrieved and target dataset while Pets has the highest difference. When crossing these values with the results in Table \ref{tab:main_table}, no correlations are observed between the EMD and obtained increase in accuracy given that the EMD for the Food dataset lies between datasets that obtain larger improvements. However, when comparing these results with Fig. \ref{fig:umap_distribution}, it is evident that the retrieved images are mostly close to the clusters formed by the images from the target dataset. 
This implies that although the Wasserstein distance between the retrieved and target datasets is low, the observed increase in accuracy may be attributed to the retrieval of images situated at the tails of the distribution, particularly notable in datasets like Food, which feature a diverse array of images.

\section{Ablations on CLIP backbones}
We perform ablations on using different CLIP backbones with our retrieved datasets $\bsetU^{\texttt{cls}}$,\ie ViT-g-14 and ViT-L-14. Consistently achieving enhanced zero-shot classification performance across these datasets, with the exception of Food. Notably, in the case of ViT-L-14, Food is already achieving an accuracy level of 97\%. The primary factor hindering improvement in this instance is that CLIP ViT-L-14 already encompasses a substantial portion of the Food dataset's distribution. Consequently, our retrieved set does not contribute additional information, see Table \ref{tab:image_class_other}.

\begin{table}[t]
\resizebox{\columnwidth}{!}{%
\begin{tabular}{rlllll}
\toprule
\multicolumn{1}{c}{\textit{}}       & Flowers  & Pets  & Cars  & Food  & ImageNet \\ \midrule
\textit{Zero Shot CLIP ViT-g-14 LAION-2B}   & 77.35    & 94.33 & 92.92 & 91.55 & 76.64    \\
Ours     & 86.53  \upg{9.18} & 94.41  \upg{0.81} & 93.73  \upg{0.81} & 91.61  \upg{0.06}  & 77.12  \upg{0.48}   \\ \cmidrule(lr){2-6}
\textit{Zero Shot CLIP ViT-L-14 OpenAI}     & 79.17    & 93.43 & 77.94 & 97.36 & 75.53     \\ 
Ours& 87.88  \upg{8.71}    & 94.28  \upg{0.85}  & 85.76  \upg{7.82}  & 97.24  \downr{0.12} & 76.06   \upg{0.53}    \\ \bottomrule
\end{tabular}
}
\caption{Performance of different CLIP backbones by using our method. We can see a consistent performance improvement in each of the datasets. }
\label{tab:image_class_other}
\end{table}

\section{Analysis of using pre-trained model vs updated model using the retrieved set}
Once the classifier is trained on the retrieved images, the decision to use either the classifier or CLIP itself during inference will be determined by Eq. (\ref{eq:prediction}), as detailed in the paper. Table \ref{tab:clipvslp_transposed} provides the proportion of instances the LP classifier is used. Datasets with the largest accuracy gains, namely Flowers and Cars, pass the largest proportion of images through the classifier. Conversely, Food101 showed minimal improvement and the lowest classifier utilization rate correspondingly. This shows the effectiveness of the entropy-driven classifier selection method: better retrieved images for Flowers and Cars led to lower entropy in the classifier's decisions, enhancing confidence in its predictions relative to CLIP. Conversely, the suboptimal image retrieval for Food101 leads to increased entropy and diminishes the reliability of classifier predictions in comparison to CLIP's outcomes.

\begin{table}[h]
    \centering
    \resizebox{\columnwidth}{!}{%
    \begin{tabular}{lccccc}
        \toprule
        & Pets & Flowers & Cars & Food101 & ImageNet \\ \midrule
        Train & 76.30 & 89.11 & 87.90 & 65.94 & 80.47 \\ 
        Test  & 75.87 & 89.44 & 87.69 & 65.95 & 80.74 \\ \bottomrule
    \end{tabular}}
    \caption{Percentage of instances from each split of a dataset passed through the classifier as opposed to CLIP according to Eq. (\ref{eq:prediction}).}
    \label{tab:clipvslp_transposed}
\end{table}

\section{Using our method for few-shot learning}
\label{app:fewshot}
\begin{figure}[h]
    \centering
    \includegraphics[height=0.93\textheight]{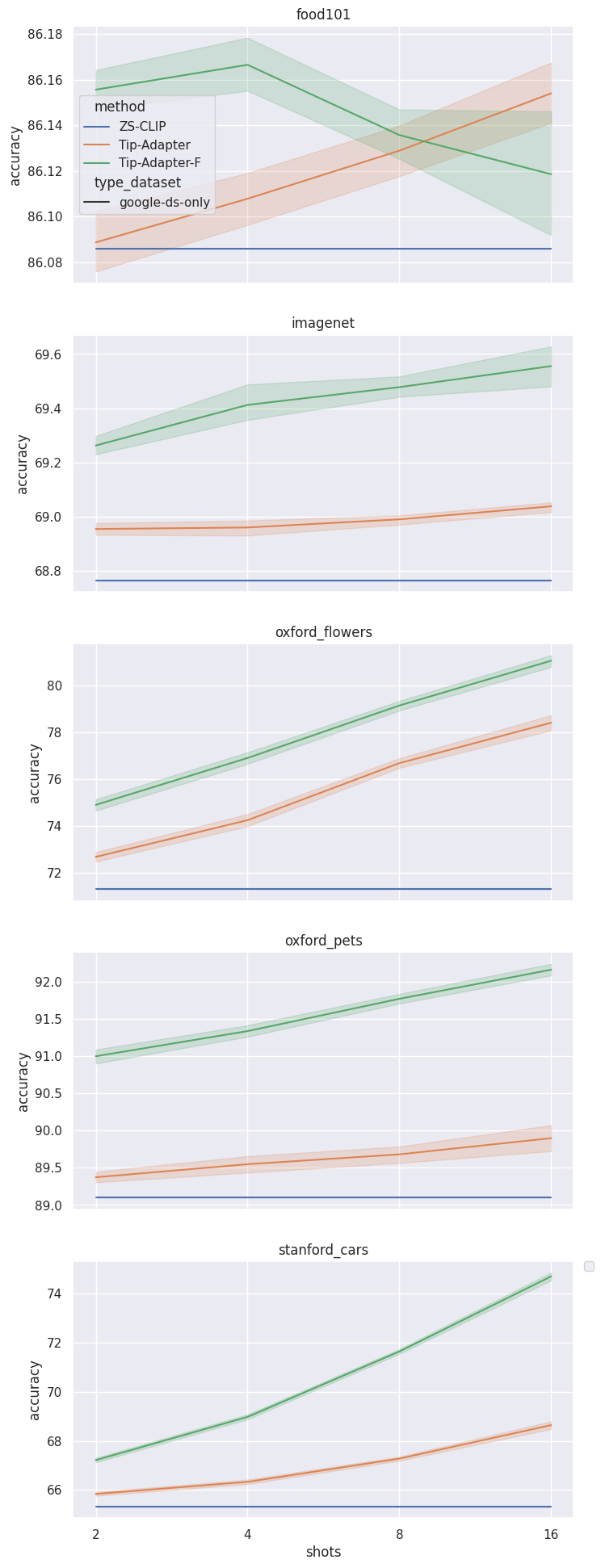}
    \caption{Few-shot experiments using Tip-Adapter \cite{zhang2021tip} with our retrieved dataset $\bsetU^{\texttt{cls}}$}
    \label{fig:fewshot}
\end{figure}
Tip-Adapter~\cite{zhang2021tip} is an adaptation technique designed for CLIP, enabling few-shot classification without extensive training. It inherits the training-free advantage observed in zero-shot CLIP, while simultaneously achieving competitive performance in comparison to methods requiring extensive training. Tip-Adapter constructs an adapter using a key-value cache model based on the few-shot training set. 
Moreover, by fine-tuning the cache model, Tip-Adapter can achieve state-of-the-art performance with significantly fewer training epochs than existing methods, demonstrating both effectiveness and efficiency. We chose this method to test our dataset since it is a training-free technique that can show how relevant it is to have a dataset that matches the distribution of the target dataset.
We performed experiments using Tip-Adapter with the retrieved dataset $\bsetU^{\texttt{cls}}$. These experiments are performed 100 times per dataset with different seeds. Thus, using different images for each k-shot. We use $k={2,4,8.16}$ instances from the retrieved dataset. Figure \ref{fig:fewshot} shows the performance in each benchmark. We noted that both variants of the Tip-Adapter (Training-Free and Fine-Tuning) derived benefits from our retrieved dataset. This underscores the effectiveness of our methodology in automatically constructing a dataset that comprehensively represents the distribution of the target dataset, extending beyond just our linear probing approach.

\section{Qualitative analysis of the refinement}
\label{app:qualitative_analysis_refinement}
\begin{figure}[h]
    \centering
    \begin{subfigure}[b]{0.475\textwidth}
        \centering
        \includegraphics[width=\textwidth]{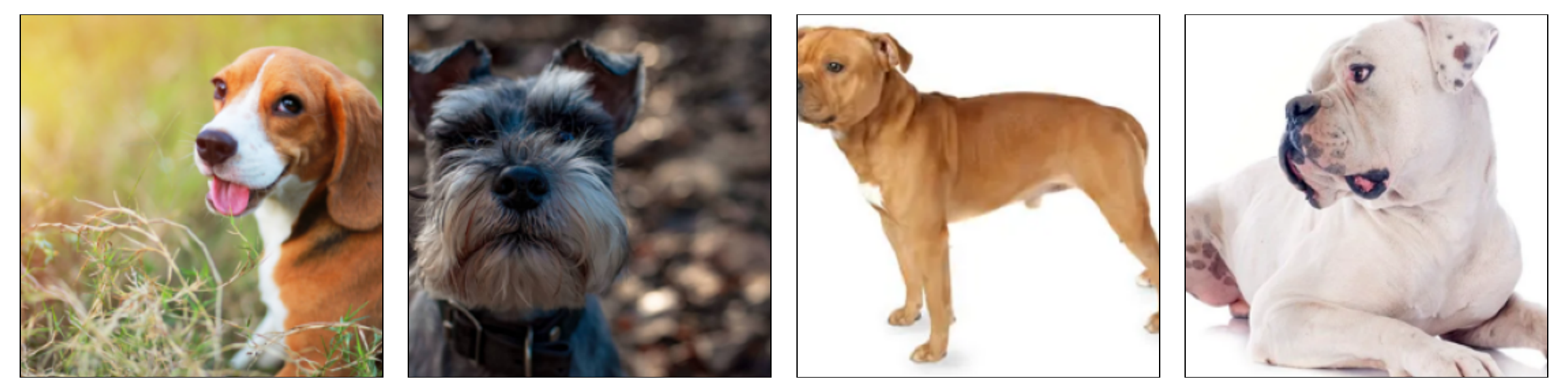}
        \caption{Accepted Samples for OxfordPetIII}
    \end{subfigure}%
    \hfill
    \begin{subfigure}[b]{0.475\textwidth}
        \centering
        \includegraphics[width=\textwidth]{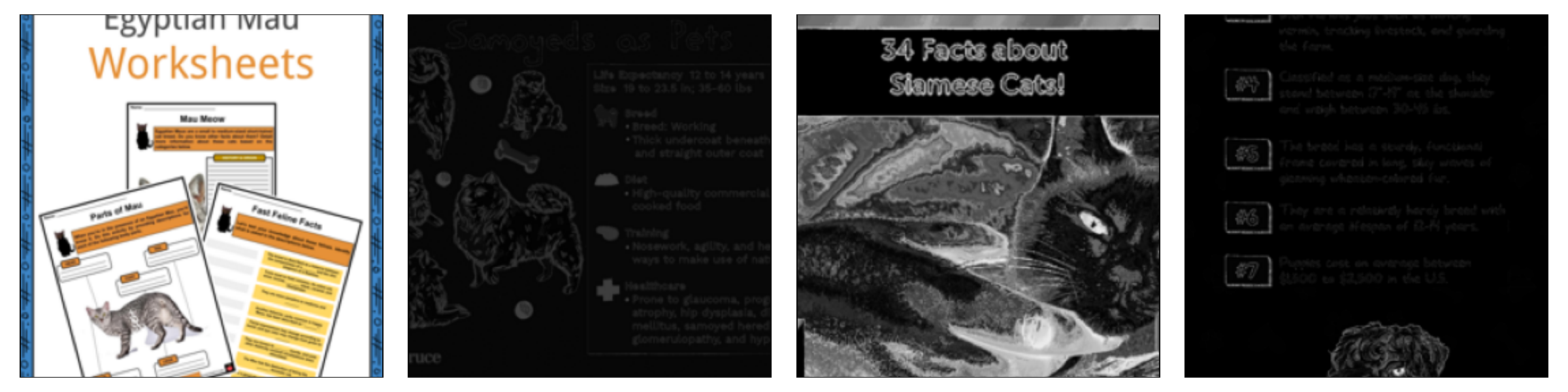}
        \caption{Rejected Samples for OxfordPetIII}
    \end{subfigure}%
    \hfill
    \begin{subfigure}[b]{0.475\textwidth}
        \centering
        \includegraphics[width=\textwidth]{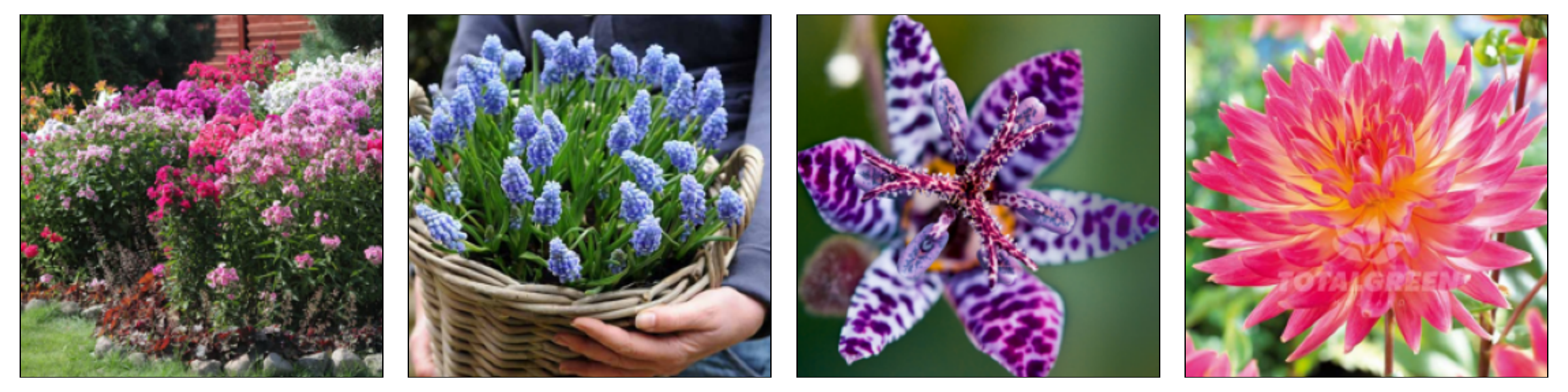}
        \caption{Accepted Samples for Flowers102}
    \end{subfigure}%
    \hfill
    \begin{subfigure}[b]{0.475\textwidth}
        \centering
        \includegraphics[width=\textwidth]{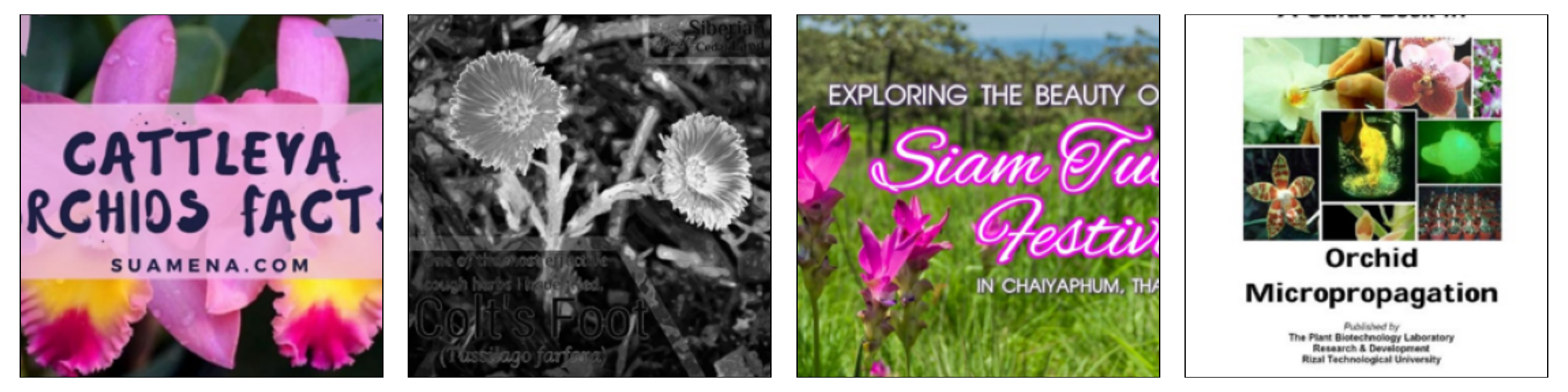}
        \caption{Rejected Samples for Flowers102}
    \end{subfigure}%
    \caption{Random samples from each retrieved dataset based on the refinement procedure described in Section \ref{sec:refinement}}
    \label{fig:refinement_samples_1}
\end{figure}

\begin{figure}[h]
    \centering
    \begin{subfigure}[b]{0.475\textwidth}
        \includegraphics[width=\textwidth]{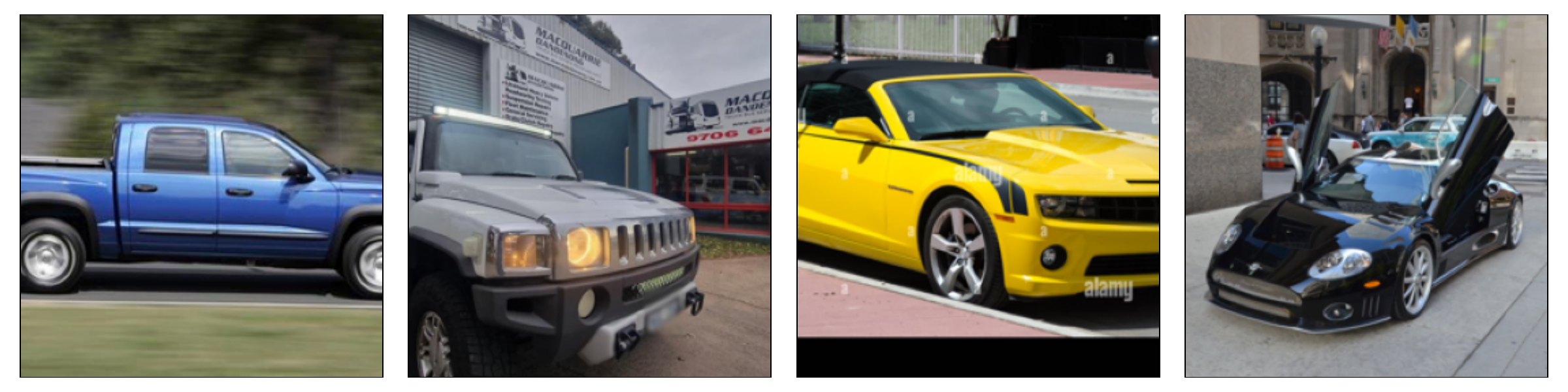}
        \caption{Accepted Samples for StanfordCars}
    \end{subfigure}%
    \hfill
    \begin{subfigure}[b]{0.475\textwidth}
        \includegraphics[width=\textwidth]{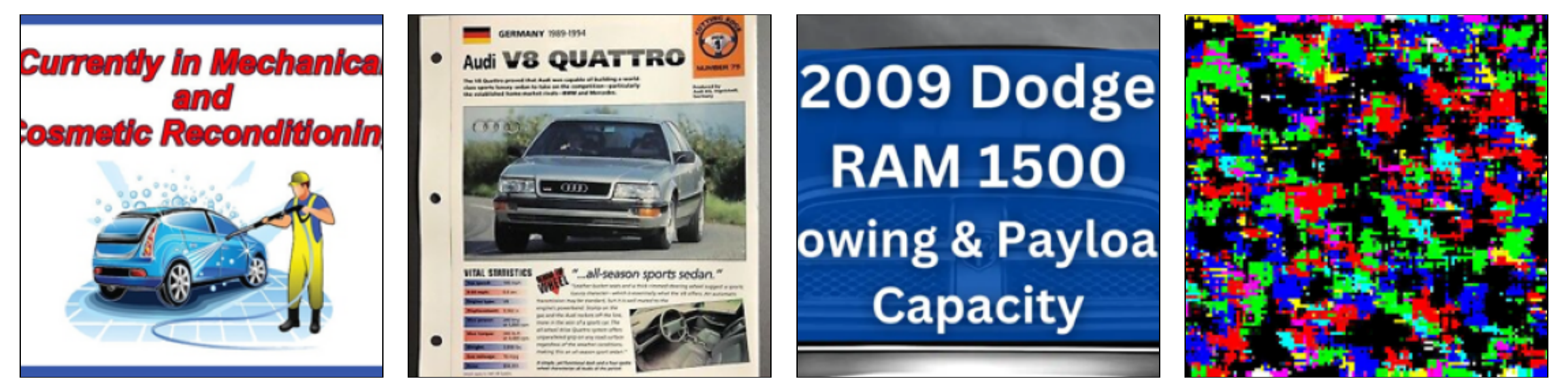}
        \caption{Rejected Samples for StanfordCars}
    \end{subfigure}%
    \hfill
    \begin{subfigure}[b]{0.475\textwidth}
        \includegraphics[width=\textwidth]{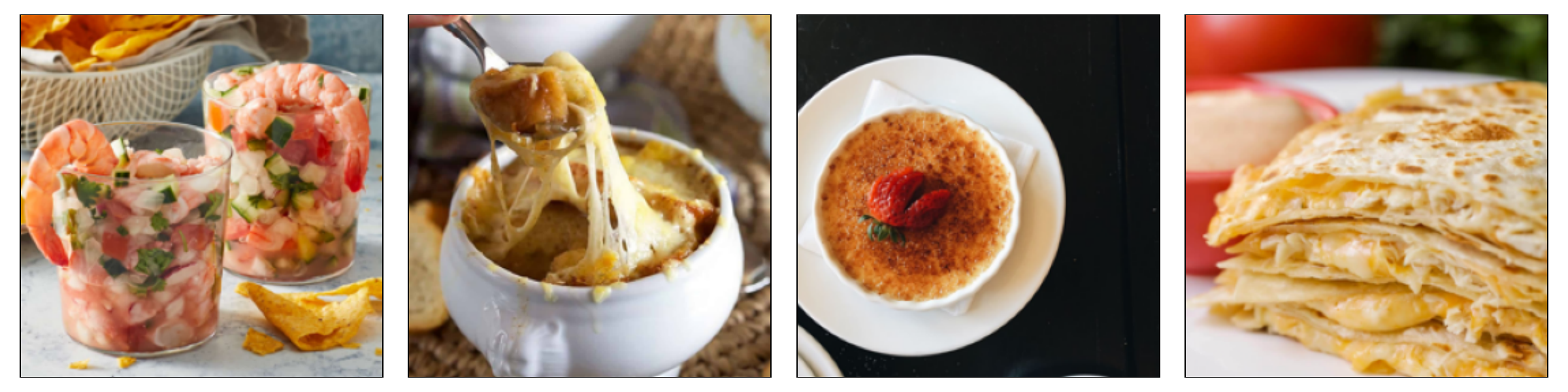}
        \caption{Accepted Samples for Food101}
    \end{subfigure}%
    \hfill
    \begin{subfigure}[b]{0.475\textwidth}
        \includegraphics[width=\textwidth]{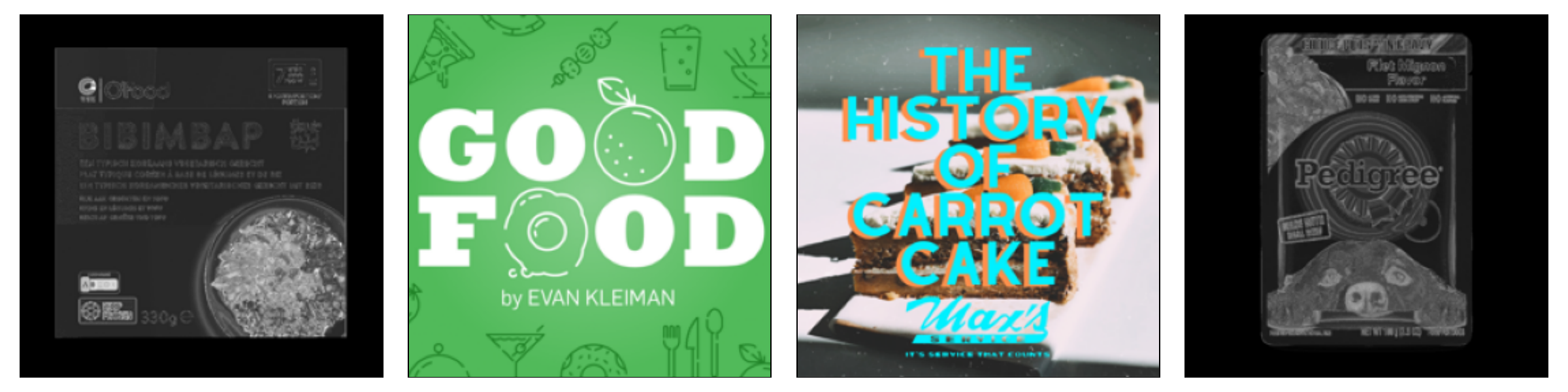}
        \caption{Rejected Samples for Food101}
    \end{subfigure}%
    \caption{Random samples from each retrieved dataset based on the refinement procedure described in Section \ref{sec:refinement}}
    \label{fig:refinement_samples_2}
\end{figure}

\begin{figure}[h]
    \centering
    \begin{subfigure}[b]{0.475\textwidth}
        \includegraphics[width=\textwidth]{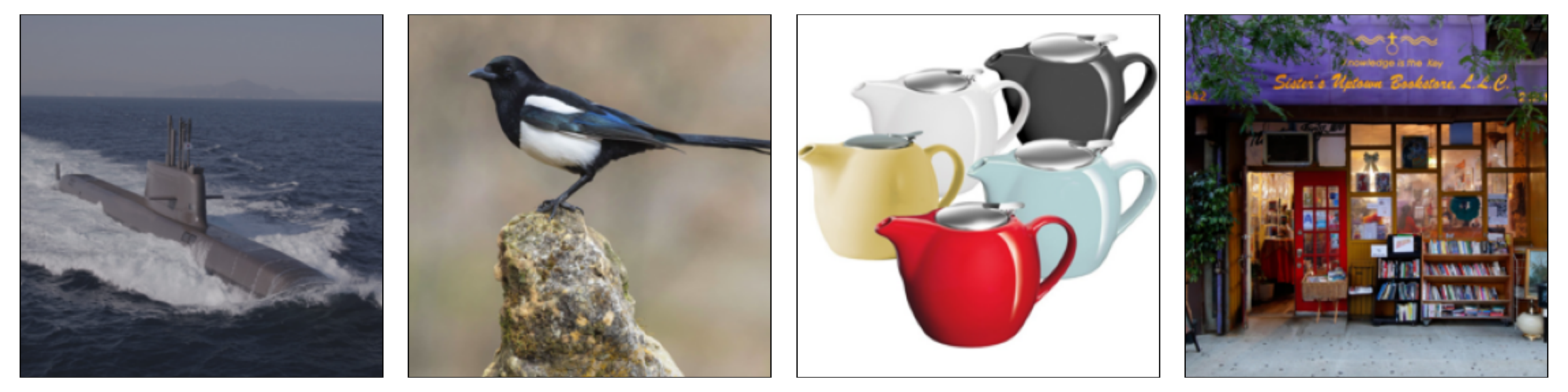}
        \caption{Accepted Samples for ImageNet}
    \end{subfigure}%
    \hfill
    \begin{subfigure}[b]{0.475\textwidth}
        \includegraphics[width=\textwidth]{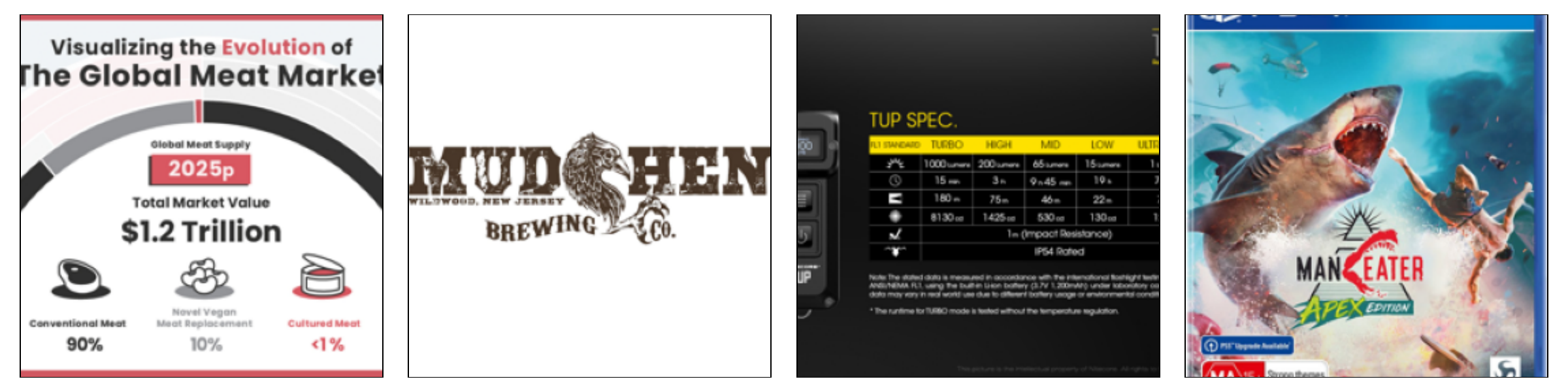}
        \caption{Rejected Samples for ImageNet}
    \end{subfigure}%
    \caption{Random samples from each retrieved dataset based on the refinement procedure described in Section \ref{sec:refinement}}
    \label{fig:refinement_samples_3}
\end{figure}

To give a more intuitive sense of what images are rejected or accepted during the refinement process we provide the visualizations in Fig. \ref{fig:refinement_samples_1}, Fig. \ref{fig:refinement_samples_2} and Fig. \ref{fig:refinement_samples_3}. These visualizations show random samples from the retrieved dataset that were accepted/rejected by the refinement procedure. A common phenomenon in the retrieval process is the acquisition of images that merely match the textual description of the queried class name but lack any shared visual similarity. As the CLIP features are employed in the refinement procedure, we effectively filter out unrelated images by applying a threshold to the similarity between the features of the retrieved and target datasets.


\section{Accuracy improvement of top classes}
\begin{figure*}
    \centering
    \begin{subfigure}{0.75\textwidth}
        \centering
        \includegraphics[width=1\linewidth]{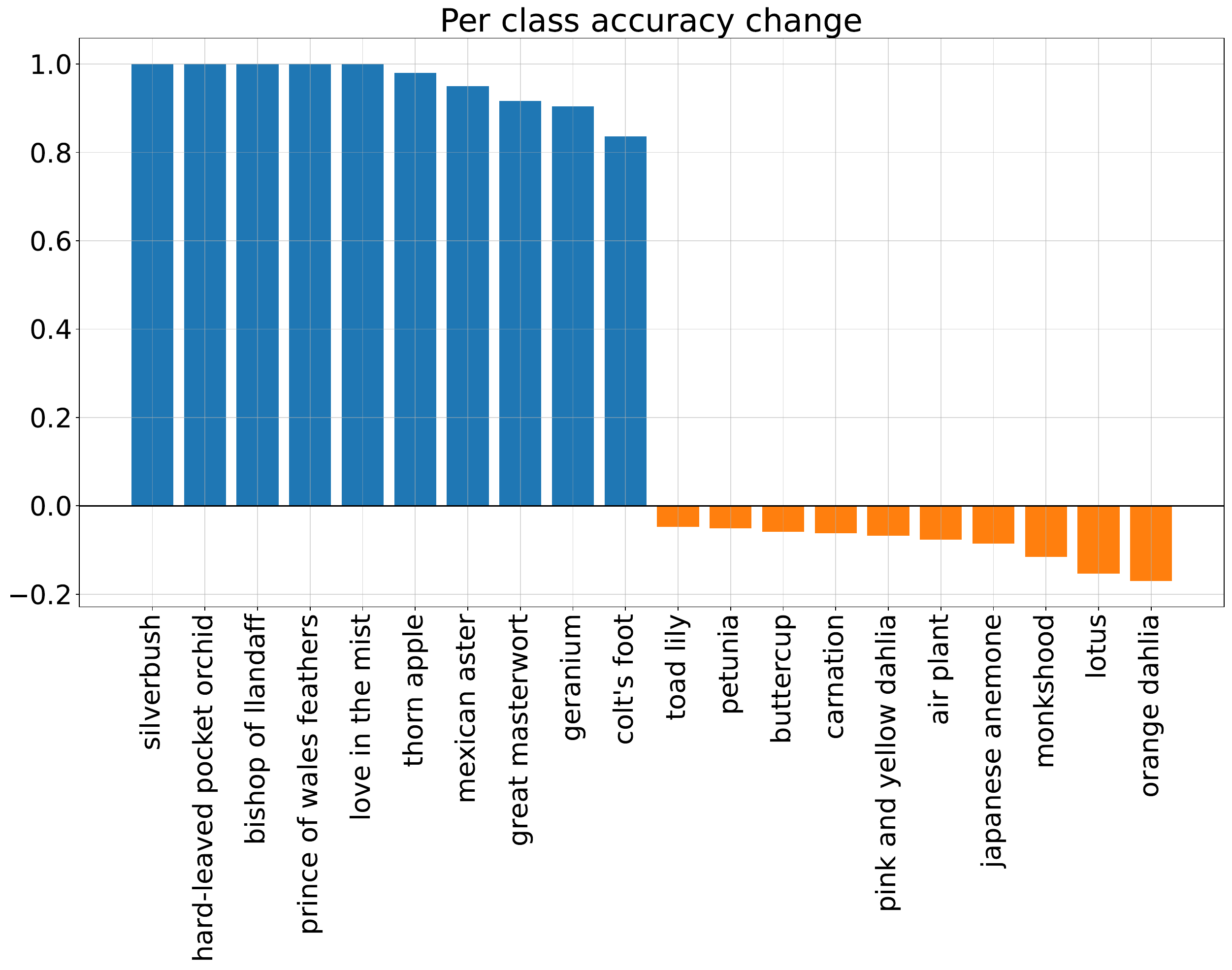}
        \caption{Flowers accuracy change for top-10 and bottom-10 classes}
        \label{fig:sub3}
    \end{subfigure}
    \hfill
    \begin{subfigure}{0.75\textwidth}
        \centering
        \includegraphics[width=1\linewidth]{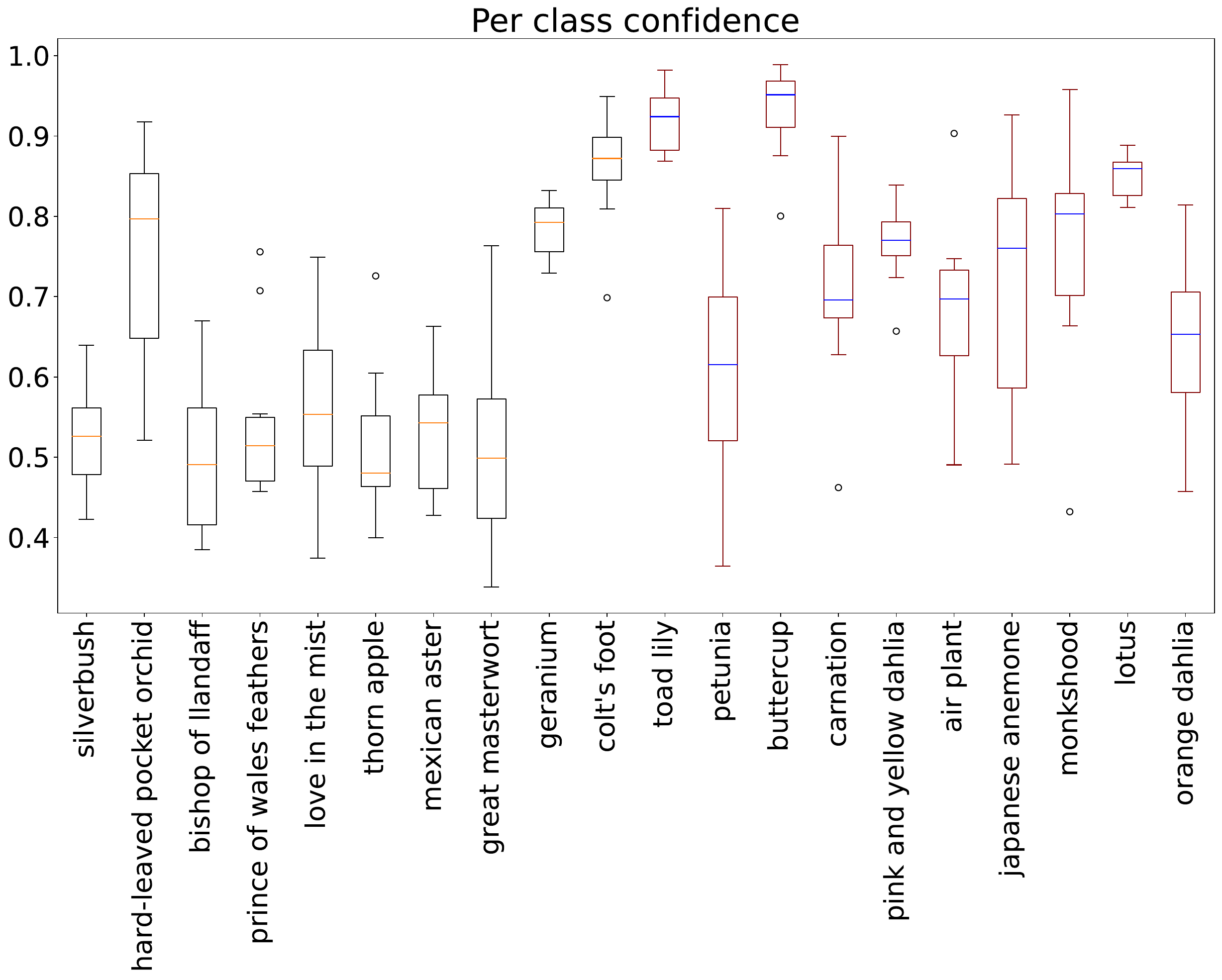}
        \caption{Flowers entropy boxplot for top-10 and bottom-10 classes}
        \label{fig:sub4}
    \end{subfigure}
    \caption{Accuracy improvements and the corresponding confidence values for each class in Flowers.}
    \label{fig:perclass_flowers}
\end{figure*}

\begin{figure*}
    \centering
    \begin{subfigure}{0.75\textwidth}
        \centering
        \includegraphics[width=1\linewidth]{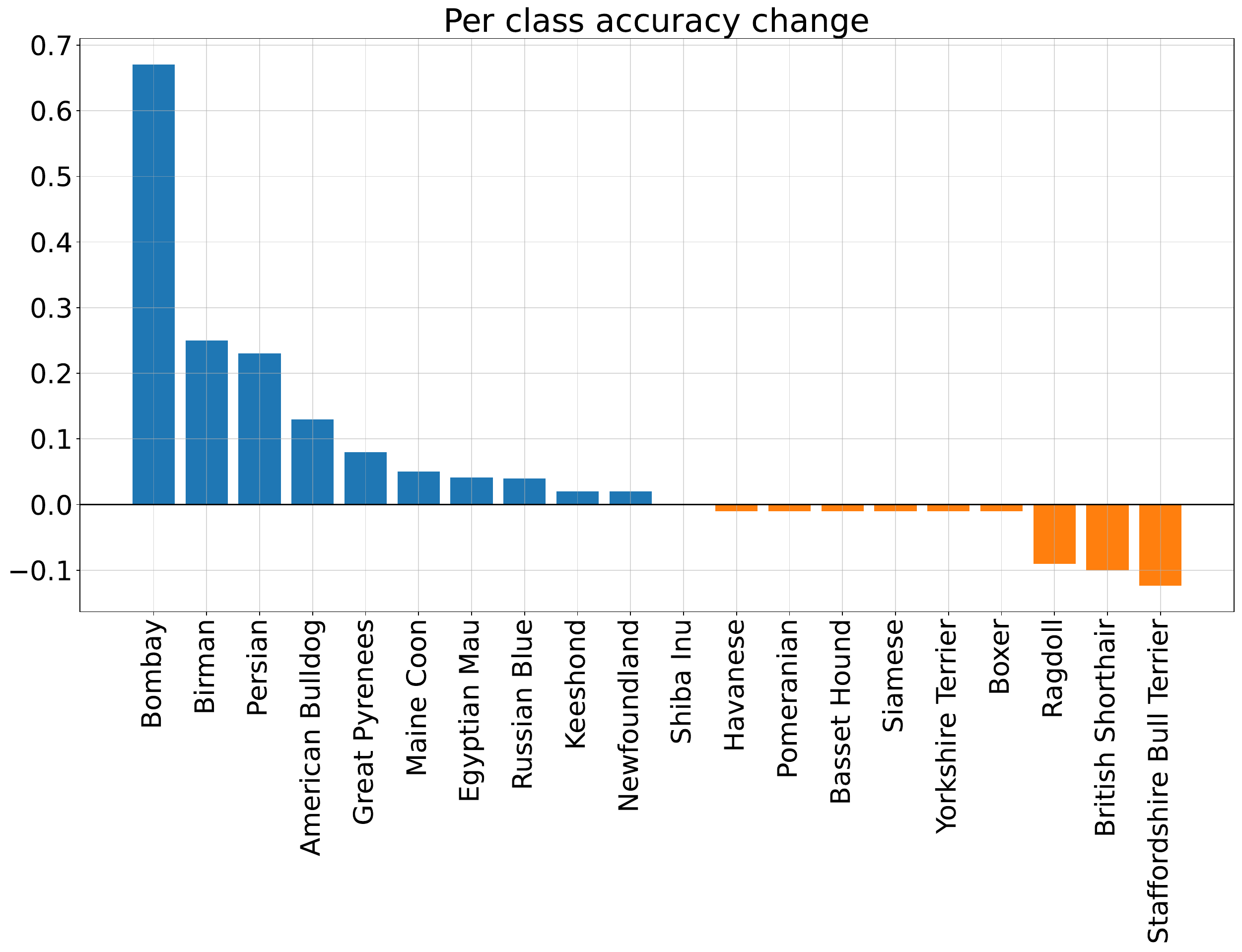}
        \caption{Pets accuracy change for top-10 and bottom-10 classes}
        \label{fig:sub1}
    \end{subfigure}
    \hfill
    \begin{subfigure}{0.75\textwidth}
        \centering
        \includegraphics[width=1\linewidth]{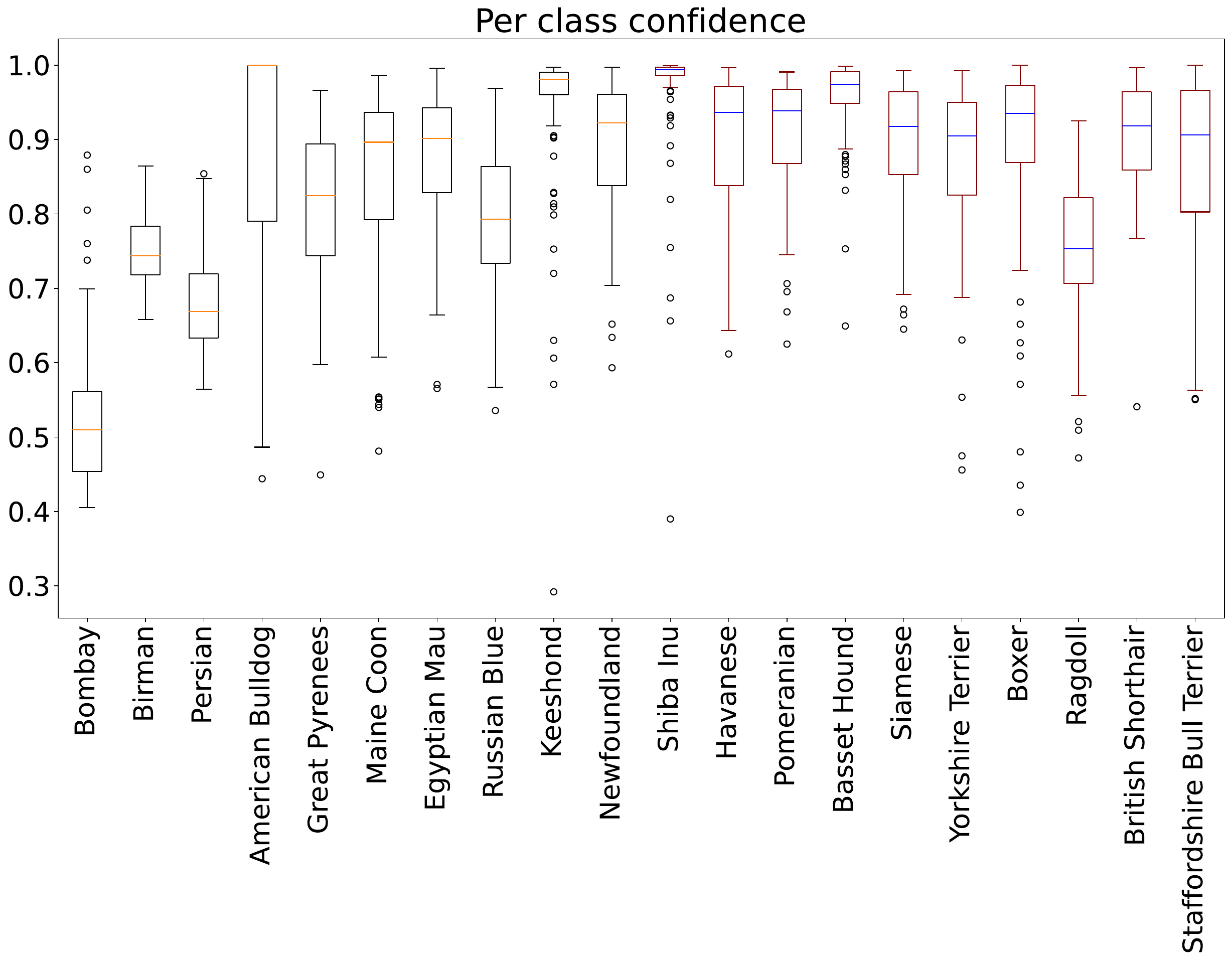}
        \caption{Pets entropy boxplot for top-10 and bottom-10 classes}
        \label{fig:sub2}
    \end{subfigure}
    \caption{Accuracy improvements and the corresponding confidence values for each class in Pets.}
    \label{fig:perclass_pets}
\end{figure*}

\begin{figure*}
    \centering
    \begin{subfigure}{0.6\textwidth}
        \centering
        \includegraphics[width=1\linewidth]{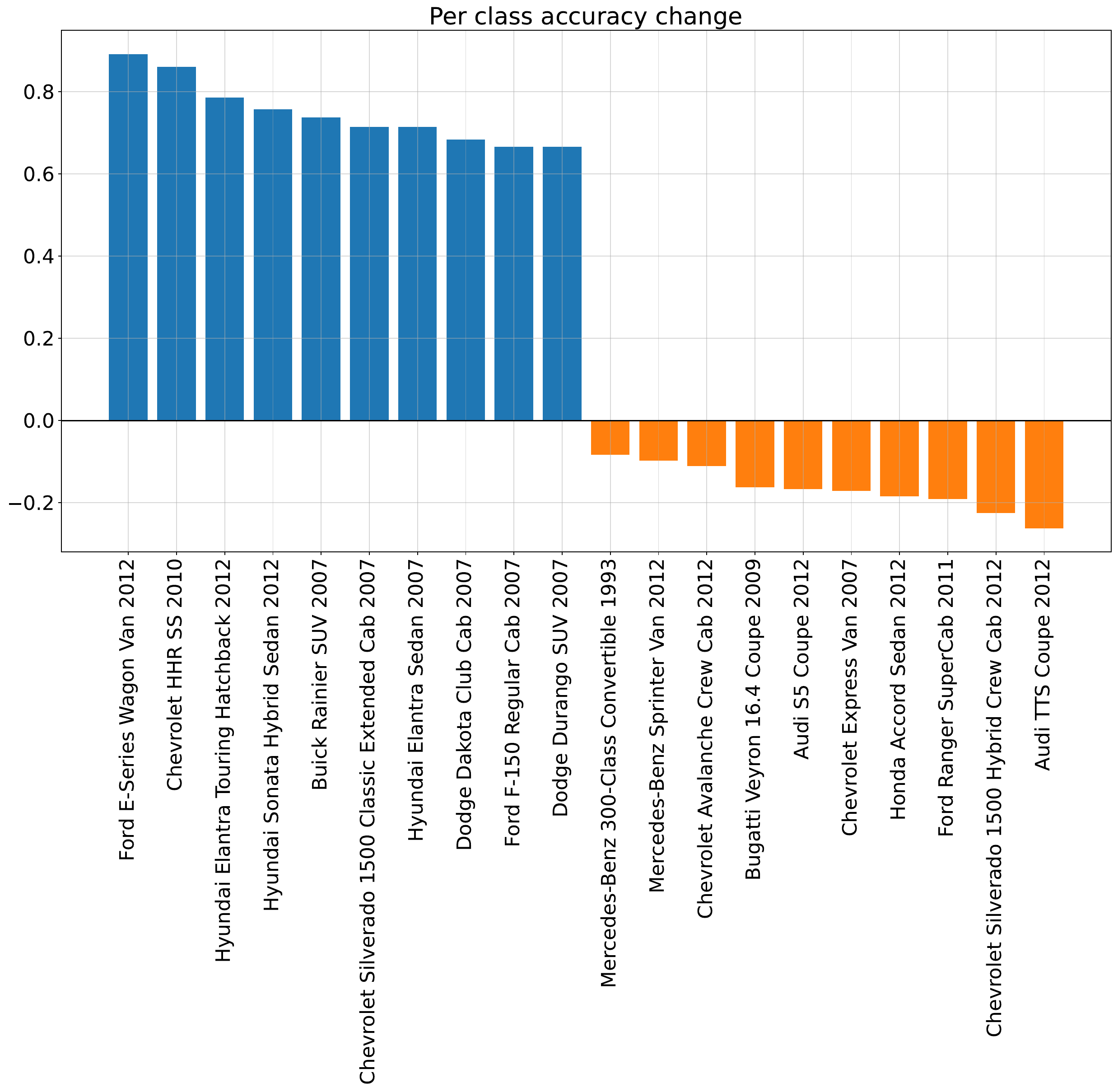}
        \caption{Cars accuracy change for top-10 and bottom-10 classes}
        \label{fig:sub3}
    \end{subfigure}
    \hfill
    \begin{subfigure}{0.6\textwidth}
        \centering
        \includegraphics[width=1\linewidth]{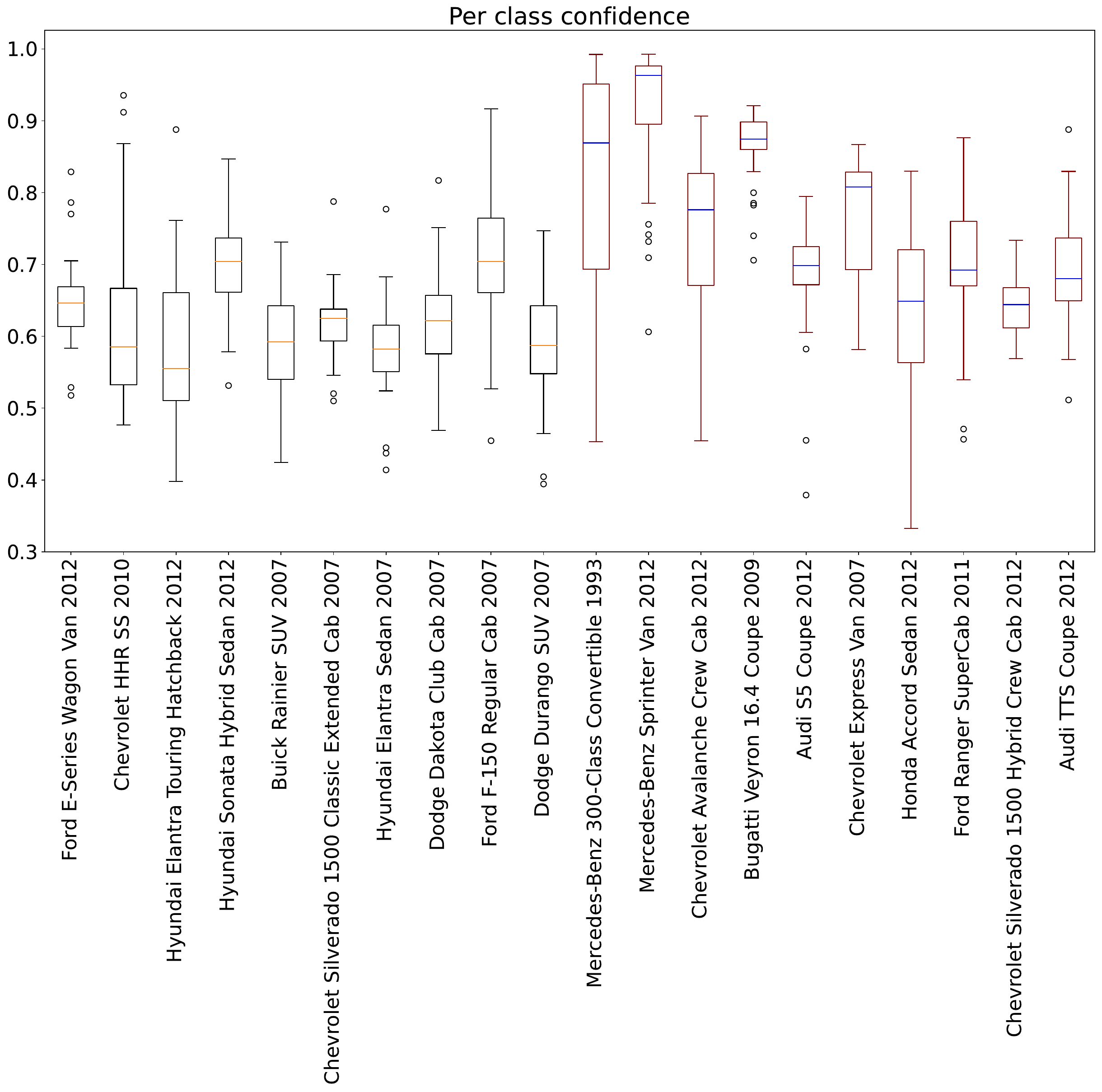}
        \caption{Cars entropy boxplot for top-10 and bottom-10 classes}
        \label{fig:sub4}
    \end{subfigure}
    \caption{Accuracy improvements and the corresponding confidence values for each class in Cars.}
    \label{fig:perclass_cars}
\end{figure*}

\begin{figure*}
    \centering
    \begin{subfigure}{0.75\textwidth}
        \centering
        \includegraphics[width=1\linewidth]{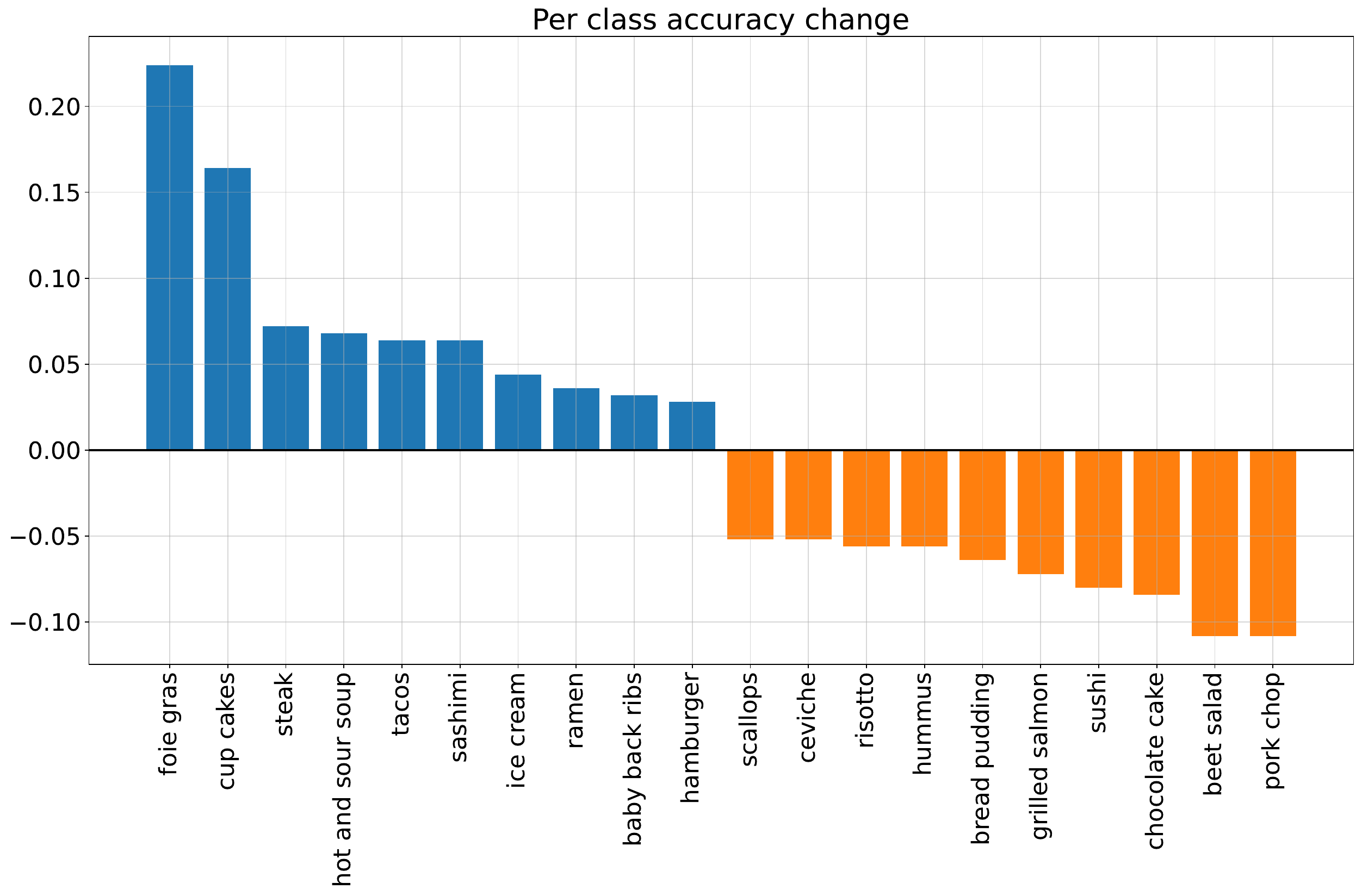}
        \caption{Food101 accuracy change for top-10 and bottom-10 classes}
        \label{fig:sub1}
    \end{subfigure}
    \hfill
    \begin{subfigure}{0.75\textwidth}
        \centering
        \includegraphics[width=1\linewidth]{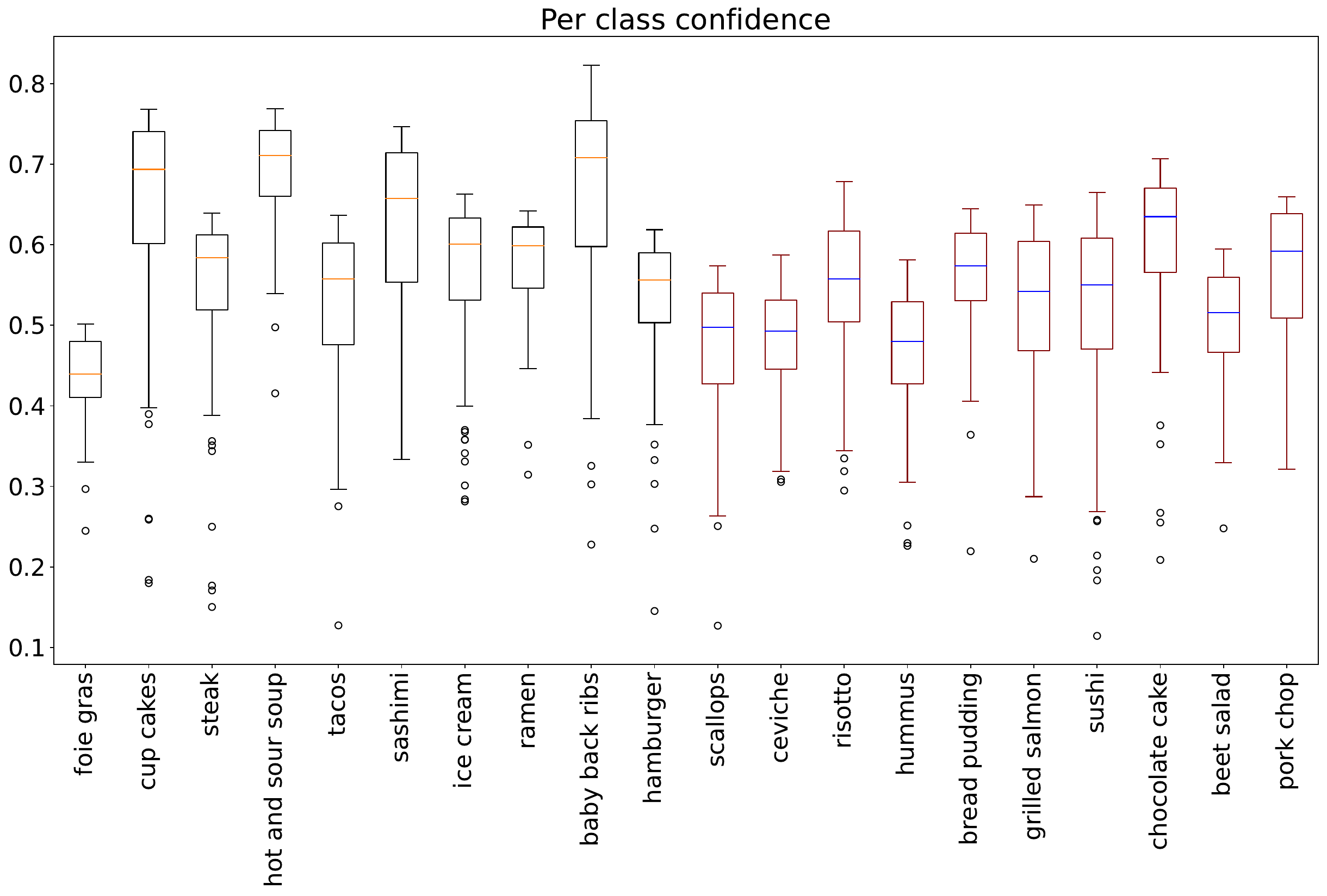}
        \caption{Food101 entropy boxplot for top-10 and bottom-10 classes}
        \label{fig:sub2}
    \end{subfigure}
    \caption{Accuracy improvements and the corresponding confidence values for each class in Food101.}
    \label{fig:perclass_food101}
\end{figure*}

\begin{figure*}
    \centering
    \begin{subfigure}{0.75\textwidth}
        \centering
        \includegraphics[width=1\linewidth]{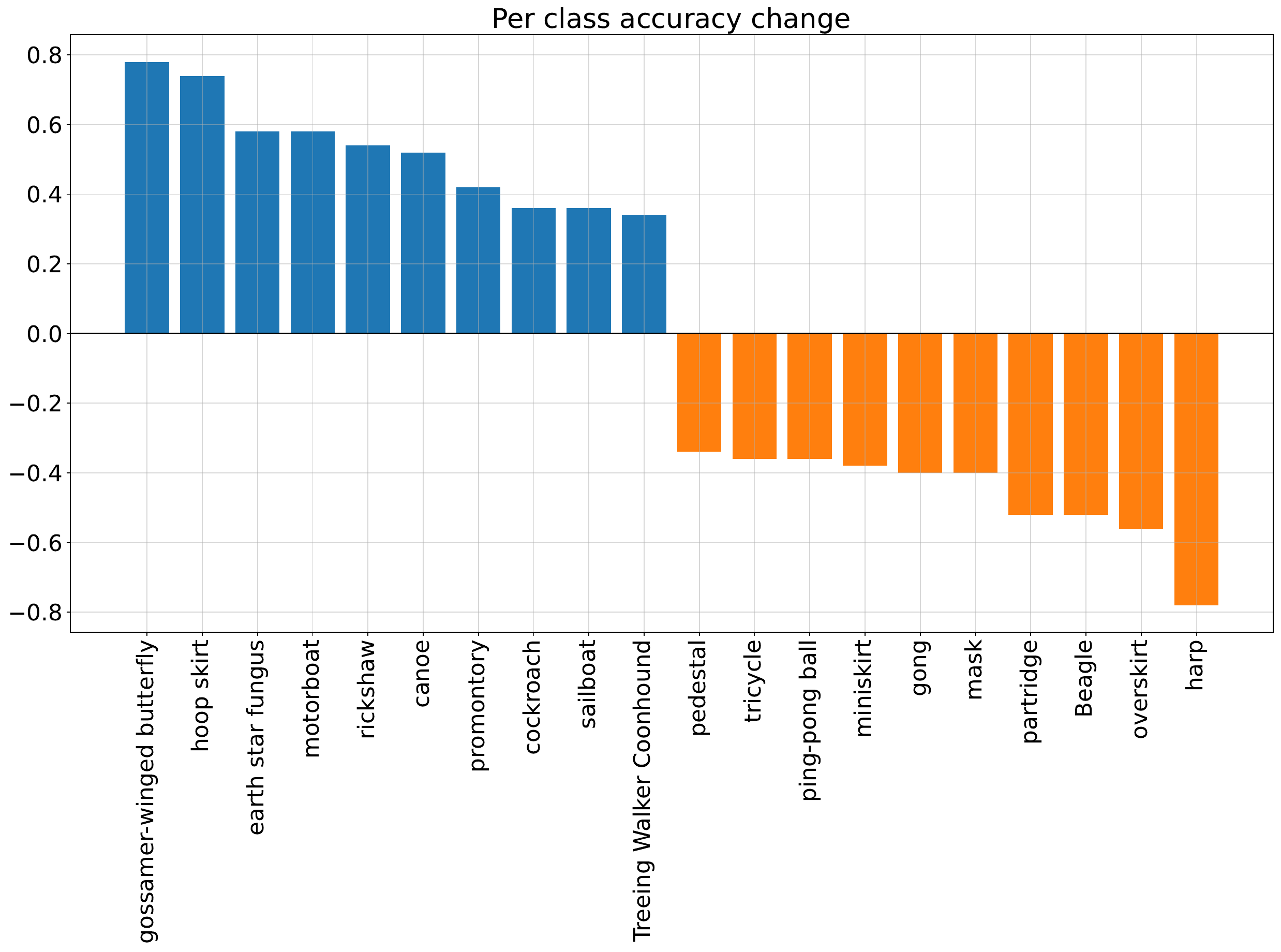}
        \caption{ImageNet accuracy changes for top-10 and bottom-10 classes}
        \label{fig:sub1}
    \end{subfigure}
    \hfill
    \begin{subfigure}{0.75\textwidth}
        \centering
        \includegraphics[width=1\linewidth]{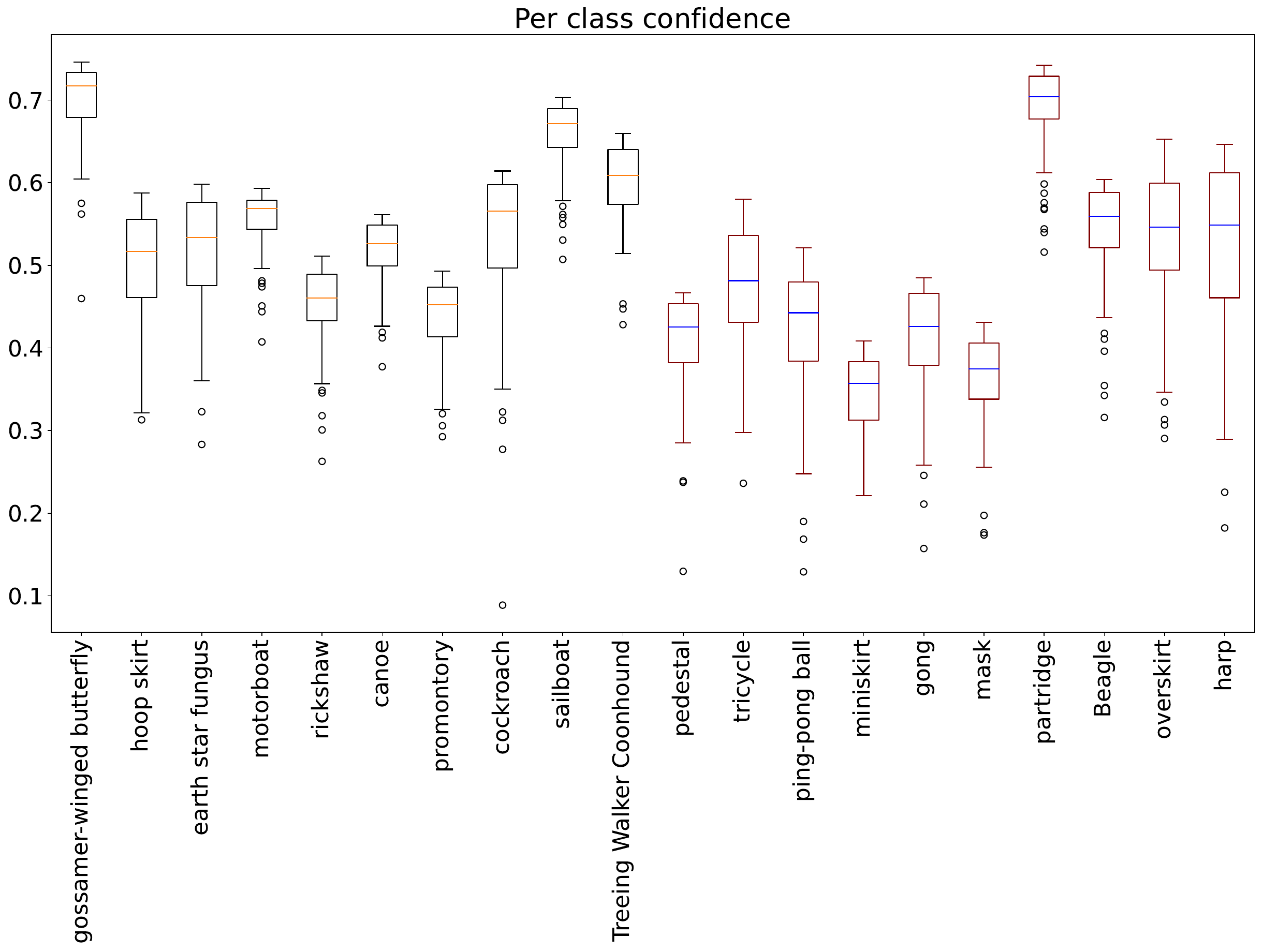}
        \caption{ImageNet confidence boxplot for top-10 and bottom-10 classes}
        \label{fig:sub2}
    \end{subfigure}
    \caption{Accuracy improvements and the corresponding confidence values for each class in ImageNet.}
    \label{fig:perclass_imagenet}
\end{figure*}

To study the changes in the accuracy of each dataset on a more granular level, we quantify the changes in the accuracy of each class. Then, we visualize the top-10 and bottom-10 classes according to these values in Fig. \ref{fig:perclass_flowers}, Fig. \ref{fig:perclass_pets}, Fig. \ref{fig:perclass_cars}, Fig. \ref{fig:perclass_food101} and Fig. \ref{fig:perclass_imagenet}. We also provide visualizations of the confidence for each of the top and bottom classes in the corresponding figure. In general, the bottom-10 classes exhibit higher confidence (lower entropy), meaning that in the case of $\bsetU^{\texttt{cls}}$, either no images are retrieved for that class or using Eq. \ref{eq:prediction}, we rely mainly on CLIP for these classes. Meanwhile, in the case of $\bsetU^{\texttt{cap}}$ due to the instance-based retrieval method that generates a query per uncertain image, more images could be retrieved for classes for which CLIP has a lower overall confidence.

\end{document}